\documentclass[acmtog]{acmart}
\usepackage{multirow}
\usepackage{enumitem}
\usepackage{colortbl}
\usepackage{makecell}
\usepackage{dsfont}
\usepackage{stfloats}

\AtBeginDocument{%
  }

\newif\ifdraft
\draftfalse

\usepackage[dvipsnames]{xcolor}

\newcommand{\revised}[1]{{\color{black}#1}}

\definecolor{gray}{rgb}{0.5,0.5,0.5}
\definecolor{green}{rgb}{0, 0.6, 0}
\definecolor{orange}{rgb}{1, 0.5, 0}
\definecolor{mahogany}{rgb}{0.75, 0.25, 0.0}
\definecolor{purple}{rgb}{0.6, 0, 0.6}
\definecolor{darkgreen}{rgb}{0, 0.3, 0}
\definecolor{orange}{rgb}{1, 0.5, 0.}
\definecolor{lightblue}{rgb}{0.52, 0.75,0.91}
\definecolor{steelblue}{rgb}{0.27, 0.51,0.71}
\definecolor{softgreen}{rgb}{0.66,0.87,0.74}
\definecolor{softred}{rgb}{0.96,0.71,0.69}

\ifdraft

  \newcommand{\rl}[1]{\textcolor{olive}{#1}}
  \newcommand{\RL}[1]{{\color{olive}{\bf RL: #1}}}
\else

  \newcommand{\rl}[1]{#1}
  \newcommand{\RL}[1]{} 
\fi

\newcommand{\ASection}[1]{\textcolor{cyan!30!blue}{#1}}

\usepackage{cleveref}
\crefname{section}{Sec.}{Secs.}
\crefname{table}{Tab.}{Tabs.} 
\crefname{figure}{Fig.}{Figs.} 
\crefname{equation}{Eq.}{Eqs.} 

\begin{document}
\setcopyright{cc}
\setcctype{by}
\acmJournal{TOG}
\acmYear{2026} \acmVolume{45} \acmNumber{4} \acmArticle{156}
\acmMonth{7} \acmDOI{10.1145/3811307}
\title{Learned Universal Interoperable Virtual Try-on}

\author{Cong Cao}
\orcid{0009-0001-5989-8367}
\email{cong.cao@mbzuai.ac.ae}
\affiliation{%
  \institution{MBZUAI}
  \city{Abu Dhabi}
  \country{United Arab Emirates}}

\author{Xianhang Cheng}
\orcid{0009-0006-9495-7330}
\email{xianhang.cheng@mbzuai.ac.ae}
\affiliation{%
  \institution{MBZUAI}
  \city{Abu Dhabi}
  \country{United Arab Emirates}}

\author{Jingyuan Liu}
\orcid{0000-0002-4648-5555}
\email{jliucb@connect.ust.hk}
\affiliation{%
  \institution{The University of Tokyo}
  \city{Tokyo}
  \country{Japan}}
\affiliation{%
  \institution{MBZUAI}
  \city{Abu Dhabi}
  \country{United Arab Emirates}}

\author{Yujian Zheng}
\orcid{0000-0001-7784-8323}
\email{yujian.zheng@mbzuai.ac.ae}
\affiliation{%
  \institution{MBZUAI}
  \city{Abu Dhabi}
  \country{United Arab Emirates}}
\authornote{Co-corresponding author}

\author{Zhenhui Lin}
\orcid{0000-0002-0482-1549}
\email{zhenhui.lin@mbzuai.ac.ae}
\affiliation{%
  \institution{MBZUAI}
  \city{Abu Dhabi}
  \country{United Arab Emirates}}

\author{Ren Li}
\orcid{0000-0003-2998-7104}
\email{ren.li@mbzuai.ac.ae}
\affiliation{%
  \institution{MBZUAI}
  \city{Abu Dhabi}
  \country{United Arab Emirates}}
\affiliation{%
  \institution{SUSTech}
  \country{China}}

\author{Meriem Chkir}
\orcid{0009-0006-0777-6364}
\email{meriem.chkir@mbzuai.ac.ae}
\affiliation{%
  \institution{MBZUAI}
  \city{Abu Dhabi}
  \country{United Arab Emirates}}

\author{Hao Li}
\orcid{0000-0002-4019-3420}
\affiliation{%
  \institution{Pinscreen}
  \country{USA}}
\affiliation{%
  \institution{MBZUAI}
  \country{United Arab Emirates}}
\email{hao@hao-li.com}
\authornotemark[1]
\renewcommand{\shortauthors}{Cao et al.}

\begin{abstract}

To enable large-scale reuse of real-world 3D assets-where garments and characters rarely share skeletons, templates, or dense correspondences-we present a fully automated virtual try-on system that dresses complex, multi-layer garments onto diverse, arbitrarily posed humanoids. Our key idea is to use SMPL as an intermediate proxy and decompose clothing-to-body transfer into two correspondence tasks with distinct challenges: (1) clothing-to-SMPL (partial-to-complete alignment) and (2) body-to-SMPL (large pose/shape variation and stylization). We address clothing-to-SMPL using a geometry-driven correspondence model, and introduce a diffusion-based body-to-SMPL correspondence approach that leverages multi-view consistent appearance features together with a pretrained 2D foundation model. Using these correspondences, we register SMPL/SMPL+D (Displacement) to the garment and target body and then perform simulator-driven fitting by transferring the garment along a smooth SMPL$\rightarrow$SMPL+D transition, producing physically plausible draping on the target. Our system handles complex garment topology (including non-manifold meshes) and generalizes to a wide range of humanoid characters (e.g., humans, robots, cartoons, and creatures) while remaining computationally practical. Upon draping, our system also supports fast customization of clothing size. We show that our system
can produce high-quality 3D clothing fittings without any human labor,
even when 2D clothing sewing patterns are not available. Our project page is: https://cao-cong0.github.io/LUIVITON-Learned-Universal-Interoperable-VIrtual-Try-ON/.


\end{abstract}

\ccsdesc[500]{Computing methodologies~Mesh models}




\keywords{Virtual Try-On, Shape Correspondence, SMPL Registration}


\begin{teaserfigure}
  \includegraphics[width=0.95\textwidth]{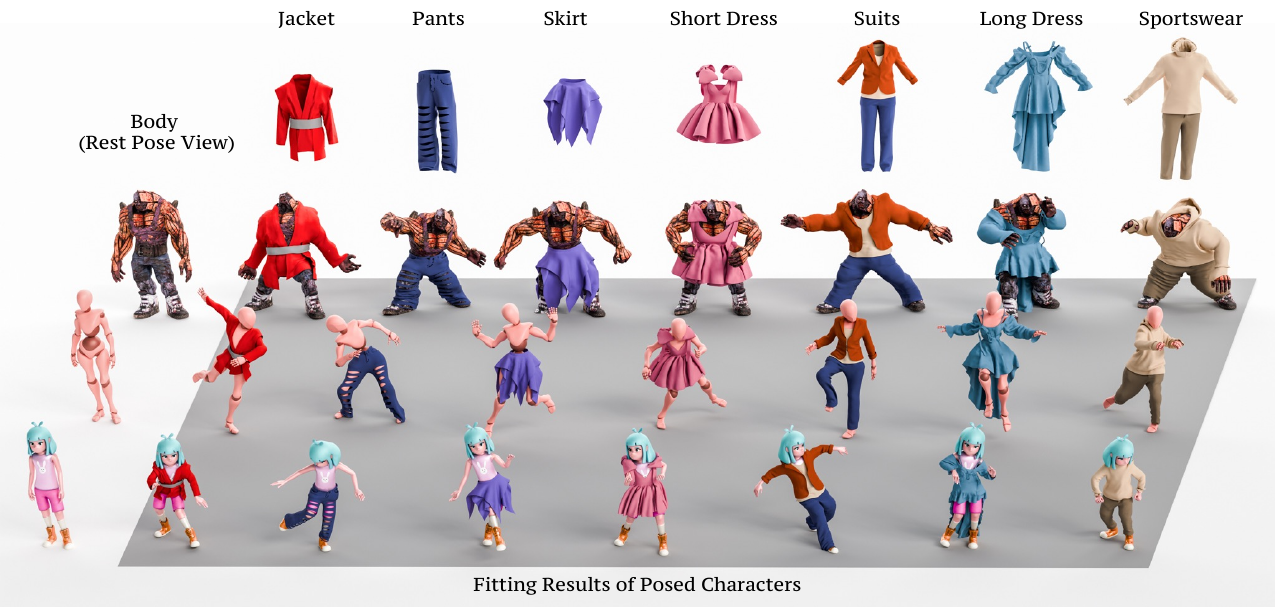}
  \centering
  \caption{We propose a fully automated and robust virtual try-on system for dressing any 3D humanoid character with arbitrary types of rest-posed 3D clothing models. Given a 3D body model and clothing (jacket, suits, pants, dress, etc.), our system automatically determines where the clothing should be placed on the body and how it should be fitted, even for unseen clothing and body, making our system suitable for virtual try-on applications. This figure displays fitting results for a diverse range of clothing on 3D characters with complex body poses and shapes. } 
  \label{fig:teaser}
\end{teaserfigure}


\maketitle

\section{Introduction}
Avatars in games, films, social media, and AR/VR applications span diverse realistic, stylized, and fantastical appearances, yet clothing remains essential for identity and visual coherence. Despite the growing availability of 3D garment assets, reusing them across characters remains costly: fitting a garment to a new body often requires extensive manual initialization, repeated simulations, and collision cleanup—especially for complex outfits.

Many existing solutions rely on assumptions that frequently fail in practice, such as a shared rig/skeleton, a standardized mannequin or parametric body, clean manifold garments, or pre-established dense correspondences. Manual tools like Marvelous Designer~\cite{marvelousdesigner} and Maya~\cite{autodesk_maya} are powerful but labor-intensive. Learning-based draping methods~\cite{patel2020tailornet, tiwari2021deepdraper, de2023drapenet, li2024isp} are typically SMPL-centric and struggle with stylized characters, large pose variation, and multilayer or non-manifold assets. Garment retargeting approaches~\cite{Intersection-Free_Garment_Retargeting, brouet2012design} assume a dressed source avatar/skeleton and often require compatible rigs or poses. In common reuse scenarios—heterogeneous asset sources, legacy datasets, or large-scale automation—one may only have an arbitrary garment mesh and an arbitrary posed humanoid body, without reliable pairings or correspondences. At the core of this problem lies a correspondence bottleneck under large pose variation and partial coverage. Garments cover only subsets of the body surface and exhibit substantial topological diversity (layers, openings, and non-manifold constructions), while targets span large pose/shape (and stylization) variation, making direct garment-to-target-body alignment ill-conditioned and error-prone.

We present a fully automated framework for dressing arbitrary 3D garments onto arbitrarily posed humanoid characters (realistic or stylized) without requiring a dressed source avatar, an animation skeleton, or predefined dense correspondences. Our key idea is to use SMPL~\cite{SMPL:2015} as an intermediate proxy and decompose dressing into two correspondence tasks: (i) clothing-SMPL (partial-to-complete correspondence) and (ii) target body-SMPL (pose/shape variation and stylization). We combine correspondence prediction with regularized registration and simulator-driven fitting to produce robust, physically plausible draping while preserving garment structure, including multilayer and non-manifold assets. \RL{"multilayer" is only mentioned in intro/conclusion, not explicitly in experiments.}

Our contributions include:
\begin{itemize}[left=0pt]
\item A universal virtual try-on framework that fully automates garment draping onto arbitrary non-parametric humanoid bodies (realistic or stylized) while handling complex geometries and multilayer stitching.
\item A partial-to-complete correspondence prediction model trained on a comprehensive dataset (300 garments with SMPL annotations) to align clothing with SMPL, ensuring robustness to non-manifold meshes and topological variations.
\item A body registration technique combining multi-view consistent diffusion features for improved spatial coherence and DINOv2 features with rich semantic priors, ensuring generalization across diverse body shapes and poses.
\item Fast customization enabling responsive clothing resizing (15 seconds per adjustment) with precomputed correspondences and registrations, streamlining artist workflows.
\end{itemize}

\section{Related Work}

\paragraph{\textbf{Garment Draping.}}
Draping aims to produce realistic cloth configurations on bodies under varying poses and shapes. Early pipelines often required careful initialization and handcrafted cues (e.g., skeleton proxies, part segmentation, or mesh partitioning) to obtain a plausible starting configuration for subsequent refinement~\cite{5246928, li2010fitting, shi2021automatic} \cite{ait20223d, huang2016automatic, wu2018fast}. More recent approaches can generate realistic drapes across body variations~\cite{patel2020tailornet, de2023drapenet, li2024isp}, but typically rely on SMPL-centric assumptions (e.g., garments initially draped on a canonical SMPL) and commonly require clean manifold garments, which limits applicability to complex garment meshes and layered outfits.

\paragraph{\textbf{Garment Retargeting and Transfer.}}
Garment retargeting transfers an existing dressed garment design from a source body to a target body while preserving design intent. Classical mesh-based transfer methods ~\cite{brouet2012design, pons2017clothcap} typically assume a source avatar/mannequin and rely on compatible structure and avatar-to-avatar correspondence. ~\cite{Intersection-Free_Garment_Retargeting} addresses the garment transfer problem by optimizing garment geometry with explicit intersection-free constraints, enabling the preservation of the original garment style, but requires the garment and target body to have similar poses and be rigged with the same skeleton topology. In contrast, we do not assume a garment-on-avatar input or dense avatar-to-avatar correspondence; instead, we use an intermediate parametric proxy (SMPL) to establish alignment and then apply simulation-based fitting, which produces physically plausible draping and pose-dependent details (e.g., wrinkles) across diverse target shapes. While our goal is not pattern-level or region-specific grading, we optionally support a global scaling of the garment rest configuration to preview overall size variants.

\paragraph{\textbf{Cloth Simulation.}}
Cloth simulation is a longstanding challenge in computer graphics, including mechanical modeling and robust collision handling. Physics-based simulation achieves high realism but is computationally expensive~\cite{ harmon2009asynchronous, li2020incremental}. Learning-based methods improve efficiency by predicting garment deformations from pose and shape~\cite{santesteban2019learning, patel2020tailornet, bertiche2020pbns, ma2020learning, santesteban2021self, bertiche2021deepsd, pan2022predicting, bertiche2022neural, santesteban2022snug}; graph-based approaches show strong generalization~\cite{grigorev2023hood}. ContourCraft~\cite{grigorev2024contourcraft} further improves robustness in challenging dressing scenarios and is therefore incorporated into our framework for simulation-based fitting, which further allows us to handle complex garment topology (including non-manifold and layered assets) commonly found in artist-created content.

\paragraph{\textbf{Shape Correspondence.}}
Shape correspondence is a fundamental challenge in computer graphics, tackled through geometric matching~\cite{besl1992method, li2009robust, li2008global, chang2010geometric, chang2012dynamic, chang2011computing}, spectral methods~\cite{jain2006robust, aflalo2016spectral}, and functional mapping~\cite{litany2017deep, ovsjanikov2012functional, ezuz2017deblurring, rodola2017partial, donati2020deep, magnet2024memory}. Most approaches are category-specific and struggle with arbitrarily dissimilar objects and non-isometric deformation. An alternative paradigm is to learn 2D features from rendered views (e.g., depth) and unproject them to 3D~\cite{wei2016dense}. With vision foundation models, 2D semantic features have improved robustness across diverse shapes~\cite{abdelreheem2023zero, dutt2024diffusion}; we follow this paradigm by incorporating synchronized multi-view diffusion features to promote spatial smoothness and completeness of correspondence.

\paragraph{\textbf{SMPL Registration.}}
SMPL~\cite{SMPL:2015} is a widely used parametric human body model defined by shape and pose parameters, and SMPL registration estimates these parameters to align the model with input 3D data. Early trivial methods implemented by ~\cite{bhatnagar2020combining, bhatnagar2020loopreg} relied on careful pose initialization provided by OpenPose ~\cite{Cao2019OpenPose} and additional cues such as texture. While segment-wise (part-based) approaches increase flexibility, they can compromise global coherence across the whole shape~\cite{zuffi2015stitched}. Recent learning-based methods, such as IP-Net~\cite{bhatnagar2020combining}, regress SMPL parameters directly but may struggle with out-of-distribution inputs. The state-of-the-art NICP~\cite{marin2025nicp} performs well on human scans but can fail on stylized humanoids due to reliance on geometry alone. Our method incorporates semantic cues from vision foundation models to improve robustness across diverse body shapes.

\begin{figure*}[htbp]
    \centering
    \includegraphics[width=0.95\linewidth]{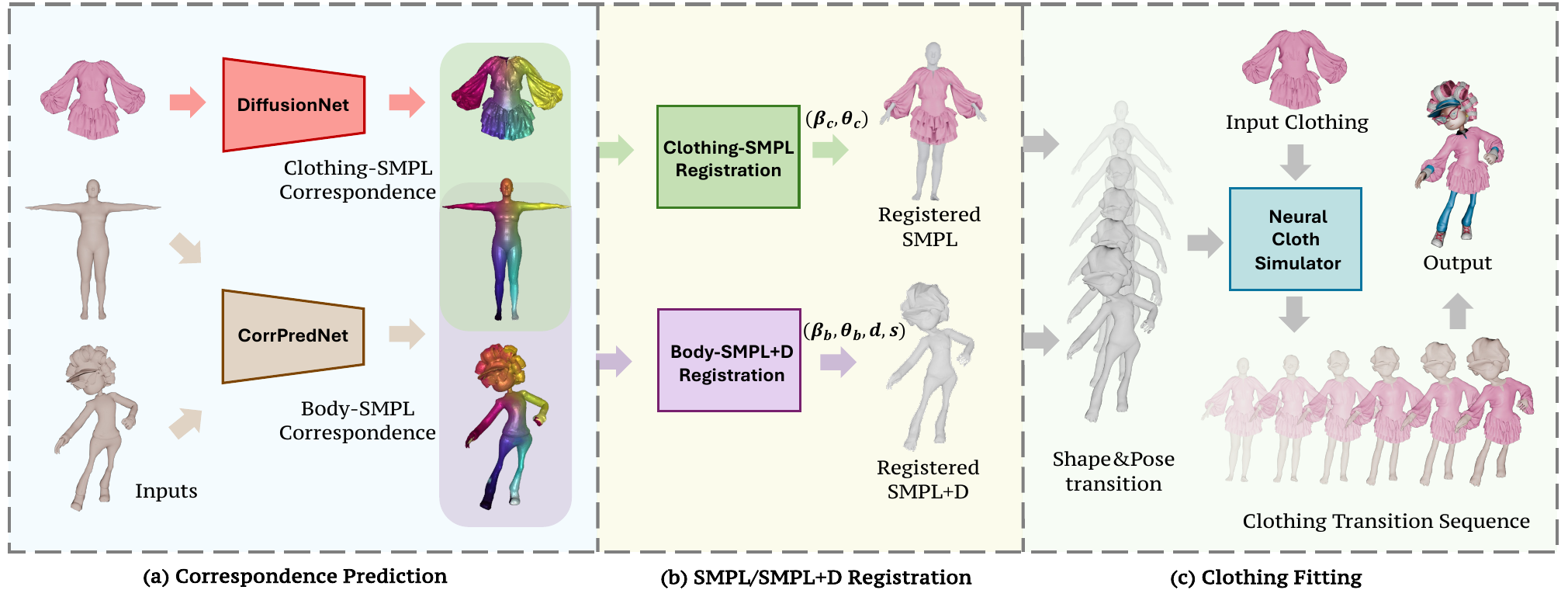}
    \caption{The overview of our system. \textbf{(a)} A modified DiffusionNet is used to compute the clothing-SMPL correspondence from the input clothing. Simultaneously, taking SMPL and the target character body as inputs, we predict the body-SMPL correspondence with our CorrPredNet. \textbf{(b)} Given the SMPL-based correspondences, we optimize the parameters of SMPL and SMPL+D by using two registration modules, which align SMPL to input clothing and body, respectively. \textbf{(c)} We perform interpolations between the registered SMPL and SMPL+D (with additional displacements and scale) to generate a smooth body sequence with shape and pose transitions. Given this sequence and clothing as input, the neural cloth simulator produces a realistic and natural fit.}
    \label{fig:overview}
\end{figure*}

\section{Method}

\cref{fig:overview} provides an overview of our system. Given a 3D garment mesh and a target humanoid body mesh, our goal is to dress the garment on the target with physically plausible draping. We achieve this by using SMPL as an intermediate proxy and executing three stages: (i) dense correspondence prediction between garment/body and SMPL (\cref{sec:clothing_corr,sec:body_corr}), (ii) body/clothing registration using these correspondences (\cref{sec:smpl_registration}), and (iii) simulator-driven clothing fitting via smooth body-shape interpolation (\cref{sec:cloth_fitting}). 

Specifically, we first estimate dense garment-SMPL and target body-SMPL correspondences; we then fit an SMPL model inside the garment and register an SMPL+D model to the target body; finally, we initialize the garment on the fitted SMPL proxy and simulate it through the SMPL$\rightarrow$SMPL+D transition until it conforms to the target body.

\subsection{Clothing-SMPL Correspondence}\label{sec:clothing_corr}

To fit an SMPL body inside a given 3D garment, it is essential to first establish consistent dense correspondences between the garment and the SMPL model. Although the recent method~\cite{dutt2024diffusion} is promising, extending it to the unexplored problem of garment-body matching is non-trivial due to the inherently partial-to-complete setting. Moreover, although garments and the human body share related semantics, this alone is insufficient for accurate correspondence prediction—semantic ambiguity, such as indistinguishable front and back regions, often leads to errors when relying solely on 2D vision priors~\cite{rombach2022high, oquab2023dinov2}. 

\begin{figure}[t]
    \centering
    \includegraphics[width=0.95\linewidth]{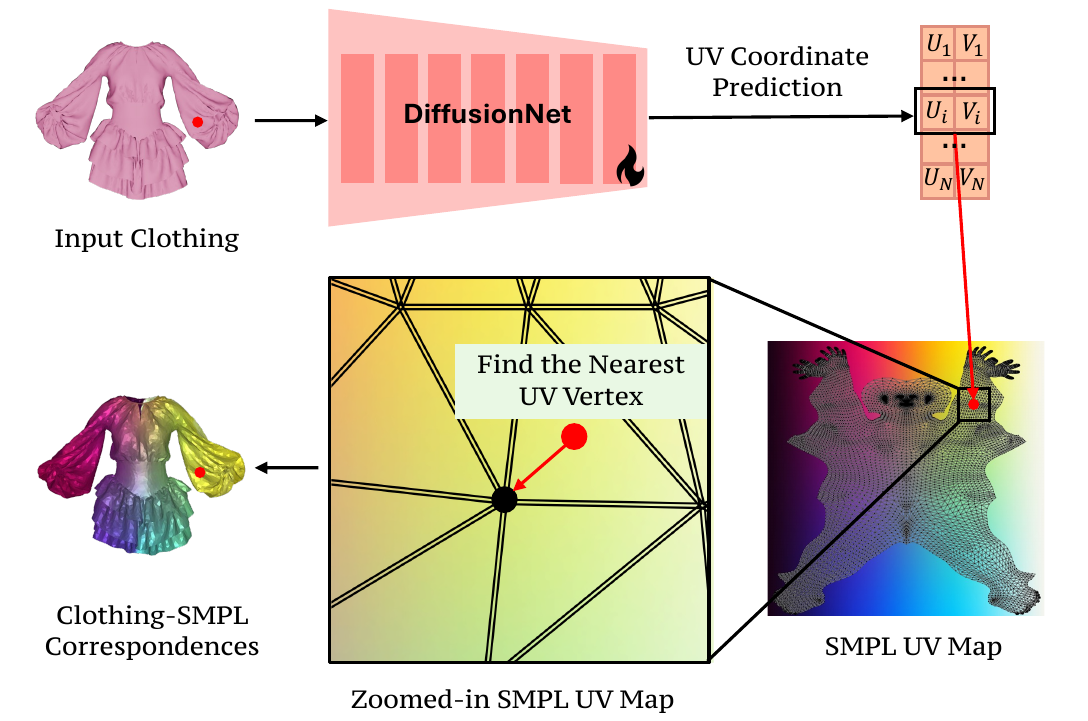}
    \caption{The clothing-SMPL correspondence module utilizes an adapted DiffusionNet to predict UV coordinates for clothing vertices, which are then used to establish correspondences between clothing and SMPL in UV space.}
    \label{fig_diffusionnet}
\end{figure}

As shown in \cref{fig_diffusionnet}, we therefore predict garment-to-SMPL correspondences by learning a supervised mapping from each garment vertex to its location in the SMPL UV space. Concretely, we learn a function \( g: \mathbb{R}^3 \to \mathbb{R}^2 \) that maps a 3D garment vertex to 2D SMPL UV coordinates, and we use DiffusionNet~\cite{sharp2022diffusionnet} as the backbone due to its effectiveness for learning on surfaces and partial-to-complete correspondence problems. Training this model requires dense ground-truth UV annotations; to this end, we build a dedicated dataset of uniquely designed garments with per-vertex SMPL UV labels for supervised learning. Additional details are provided in \ASection{Suppl. A.1} of the supplementary material.

DiffusionNet takes the garment mesh as input and predicts the UV coordinate at each vertex, treating the 2D UV coordinates on the SMPL body as surface features of the garment. The network is composed of repeated blocks that perform feature diffusion over a learned time scale, extract spatial gradients, and apply a shared pointwise MLP at each vertex. We train the network with the loss:
\begin{equation}
\begin{split}
    \mathcal{L}_{dif} = \sum_{i=1}^{N} \lVert x_i - g(X_i)\rVert,
\end{split}
\end{equation}
where $X_i$ and $x_i$ denote the 3D garment vertex and its corresponding 2D position on the SMPL UV map, respectively, and $N$ is the number of garment vertices.

\subsection{Body-SMPL Correspondence}\label{sec:body_corr}

\begin{figure*}[!th]
    \centering
    \includegraphics[width=0.95\linewidth]{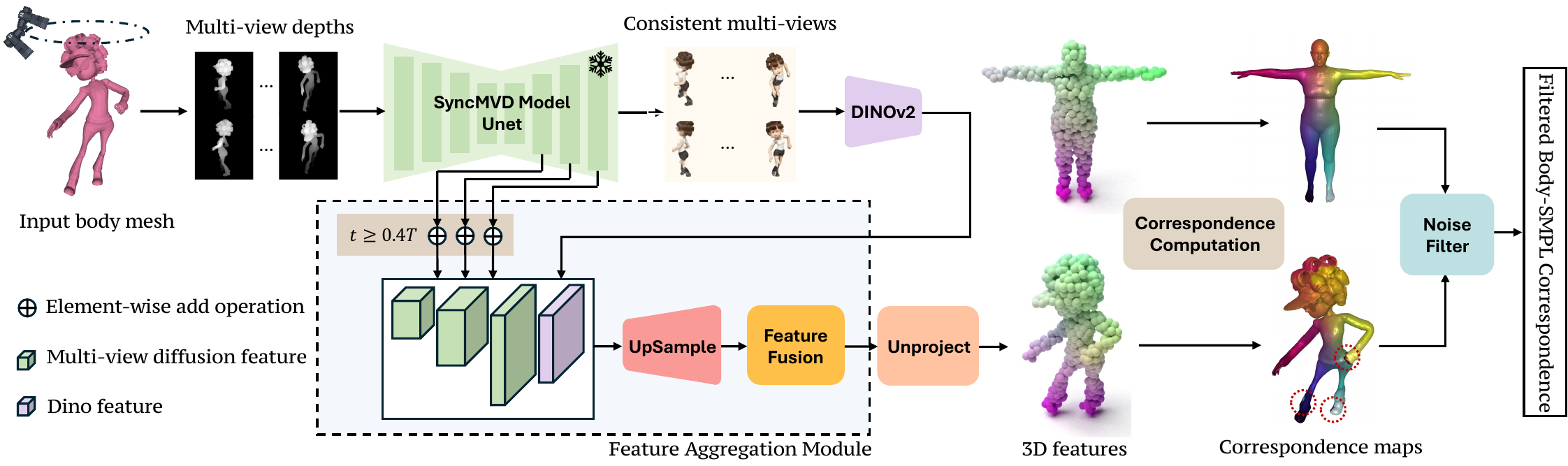}
    \caption{Starting from a body mesh, we first render multi-view depth maps and feed them into SyncMVD to generate consistent multi-view images. Next, a feature aggregation module fuses diffusion features from the SyncMVD U-Net with semantic features from DINOv2. These aggregated features are then unprojected onto the input mesh to form 3D features, drive the computation of the correspondence. Features are extracted for both the SMPL and the target body in this way. Finally, a noise filtering module is applied to identify the noise (highlighted by red circle) and produce clean Body-SMPL correspondences.}
    \label{fig:corr_predictionnet}
\end{figure*}

The body-SMPL correspondence maps vertices from a 3D humanoid body to the SMPL model, as shown in \cref{fig:corr_predictionnet}. Unlike clothing-SMPL correspondence, 
body-SMPL correspondence faces challenges due to extreme proportions and diverse poses, making geometric features less effective. Instead, semantic features offer a more robust solution.

We introduce the Correspondence Prediction Network (CorrPredNet), a diffusion model-based approach that leverages semantic features from 2D vision foundation models to extract features of both canonical SMPL body and target body for correspondence establishment. Specifically, we input multi-view depth images into a synchronized diffusion model and DINOv2~\cite{oquab2023dinov2} to obtain dense per-pixel semantic features. Following prior works~\cite{wei2016dense,dutt2024diffusion}, these 2D features are unprojected into 3D using known camera parameters, yielding per-vertex feature embeddings for both the input body and the SMPL model.
Correspondence is then established by computing cosine similarity between the two sets of 3D features. To improve robustness, we introduce a noise filtering module that removes outlier correspondences before registration.

\paragraph{\textbf{View-Consistent Feature Extraction.}} 
As shown in \cref{fig:corr_predictionnet}, we begin by rendering multi-view depth maps of the input 3D mesh, which are then passed to the SyncMVD~\cite{liu2024text} model to synthesize consistent and semantically plausible multi-view textures \revised{using a fixed text prompt input "\textit{A photorealistic human with T-shirt and shorts.}". We use depth maps to remain applicable to untextured meshes (e.g., SMPL).} 
During texture generation, we extract diffusion features from the last three layers of the UNet to capture rich and comprehensive semantic information across multiple feature scales. Specifically, for each view, the features are extracted during the denoising process for steps \( t \geq 0.4T \), where \( T \) is the total number of denoising steps. To fuse diffusion features across different time steps, we upsample all the features to the same resolution and adopt a weighted fusion strategy defined as follows:

\begin{equation}
    f^{(n)}_{\text{diff}} = \sum_{t=0.4T}^{T} f_{\text{diff},t}^{(n)} \cdot w_{t}, \quad n \in \{0, 1, 2\}
\end{equation}

\begin{equation}
    w_{t} = \frac{t - 0.4T}{T - 0.4T} \cdot (1 - w_{\min}) + w_{\min},
\end{equation}
where $f_{\text{diff},t}^{(n)}$ denotes the feature extracted at denoising step $t$ from UNet layer $n$, and \( w_t \) is the corresponding weight. This schedule biases the fusion toward features from the noisier portion of the denoising trajectory, which tend to encode more global, high-level semantics and coarse structural layout, while still retaining contributions from cleaner steps for local detail and spatial precision. We empirically set $w_{\min}=0.1$ in our method.


The consistent multi-view images of the textured 3D mesh generated by SyncMVD ~\cite{liu2024text} are sent to DINOv2 ~\cite{oquab2023dinov2} to extract semantic features. We propose a Feature Aggregation Module that produces the fused representation $f_{\text{fuse}}$ to integrate diffusion and semantic features. Details of this module are provided in the \ASection{Suppl. B.2}. 




\rl{While cosine similarity computed on the fused features $f_{fuse}$ of the SMPL and the target body provides initial correspondences,}
alignment errors, particularly in the limbs, persist due to the left-right ambiguity in the diffusion and semantic features. These errors can result in inaccurate correspondences and ultimately compromise the quality of SMPL registration. To address this issue, we introduce a noise filter module.


\paragraph{\textbf{Noise Filter.}}  

We propose an energy-based iterative noise filter that removes unreliable correspondences based on local shape differences, thereby improving the robustness of subsequent registration.

For each input body vertex \(v_i\), we define its \(K\)-ring neighborhood \(\mathcal{N}_i\). For every neighbor vertex \({v}_j \in \mathcal{N}_i\), we compute the relative difference vectors \(p_j\) and \(q_j\) for every neighbor of vertex \({v}_i\) as:
\begin{equation}
    p_j = v_j - v_i \quad \text{and} \quad q_j = \tilde{v}_j - \tilde{v}_i,
\end{equation}
where \(v_i\) and \(v_j\) are vertices from the source mesh, \(v_j\) is a neighbor of \(v_i\), and \(\tilde{v}_i\) and \(\tilde{v}_j\) are their corresponding vertices in SMPL. These vectors capture both distance and orientation differences while factoring out translations.

The optimal rotation matrix \(R_i\) is determined via Singular Value Decomposition of the covariance matrix constructed from the \(p_j\) and \(q_j\). The deformation energy for vertex \(i\) is defined as:
\begin{equation}
    E_i = \sum_{j \in \mathcal{N}_i} \|q_j - R_i p_j\|^2.
\end{equation}
This energy measures local misalignment based on contributions from all \(K\)-ring neighbors. At each iteration, we compute \(E_i\) for all vertices. A dynamic threshold \(\epsilon = \mu_E + 0.5\,\sigma_E\) is set to filter the vertices with higher energy, where \(\mu_E\) and \(\sigma_E\) are the mean and standard deviation of the energies. And for each iteration, the neighborhood size is dynamically increased as \(K = \text{iter}^2,\)
with \(\text{iter}\) representing the current iteration index, progressively incorporating a broader local context. This iterative process continues until the number of valid correspondences stabilizes or a maximum of four iterations is reached, ensuring robust noise filtering. By penalizing local misalignments, the filter helps resolve left-right ambiguities, thus enhancing SMPL registration accuracy.

\subsection{Body/Clothing Registration}
\label{sec:smpl_registration}

As shown in \cref{fig:overview}(b), the goal of this stage is to construct a shared SMPL-based proxy that connects the input garment and the target body. Using the dense correspondences estimated to SMPL in the previous stage, we (1) fit an SMPL model inside the input garment, and (2) register an SMPL+D model to the target body. We formulate these two steps as the following optimization problems. All loss definitions are detailed in \ASection{Suppl. B.5}.

\paragraph{\textbf{Clothing-SMPL Registration.}}\label{C-S registration}  
We optimize the SMPL parameters \((\theta_{c}, \beta_{c})\) to align the SMPL within the input garment. The registration is guided by the following weighted loss function:

\begin{equation} \label{smpl}
\begin{split}
    \mathcal{L} = \lambda_{c2s} \mathcal{L}_{c2s} + \lambda_{shape} \mathcal{L}_{shape} + \lambda_{pose} \mathcal{L}_{pose}  \\ + \lambda_{lap} \mathcal{L}_{lap} + \lambda_{pene} \mathcal{L}_{pene},
\end{split}
 \end{equation}  
where \(\lambda\) terms control the weighting of each loss. Specifically, \(\mathcal{L}_{c2s}\) enforces alignment between clothing and SMPL vertices, while \(\mathcal{L}_{shape}\) and \(\mathcal{L}_{pose}\) regularize shape and pose to prevent extreme unrealistic deformations. \(\mathcal{L}_{lap}\) ensures smooth deformations by Laplacian regularization, and \(\mathcal{L}_{pene}\) penalizes body-clothing interpenetration.
This step ensures an accurate SMPL fit to the clothing while maintaining realism and preventing penetration.

\paragraph{\textbf{Body-SMPL+D Registration.}}  
We optimize the SMPL+D parameters \((\theta_{b}, \beta_{b}, d, s)\), where \( d \) allows fine per-vertex displacement, and \( s \in \mathbb{R}^3 \) controls scaling along the \( x, y, z \) axes. Scaling is essential for adapting SMPL to characters with extreme proportions when minor displacements are insufficient.  

The optimization minimizes the following objective:

\begin{equation} \label{smpld}
\begin{split}
\mathcal{L} = \lambda_{b2s} \mathcal{L}_{b2s} + \lambda_{s2b} \mathcal{L}_{s2b}  + \lambda_{corres} \mathcal{L}_{corres}+ \\ \lambda_{shape} \mathcal{L}_{shape}  + \lambda_{d} \mathcal{L}_{d}  + \lambda_{s} \mathcal{L}_{s}  + \lambda_{lap} \mathcal{L}_{lap},
\end{split}
\end{equation}  
where $\mathcal{L}_{b2s}$ and $\mathcal{L}_{s2b}$ are the bidirectional point-to-mesh distances between the SMPL surface and the input body mesh, while \(\mathcal{L}_{corres}\) minimizes the distance between corresponding vertices with noise-filtered correspondence. \(\mathcal{L}_{s}\) regularizes scaling to avoid extreme unrealistic deformations, and \(\mathcal{L}_{d}\) constrains displacements for natural alignment. By jointly optimizing these terms, our method achieves robust and accurate SMPL+D registration across a wide range of body shapes and poses.

\subsection{Clothing Fitting}
\label{sec:cloth_fitting}

In the Clothing Fitting stage (\cref{fig:overview}(c)), our goal is to drape the registered garment onto the input humanoid body while preserving its original geometry and design details. Specifically, we transfer the clothing from the fitted source body (SMPL) inside the garment to the target body representation (SMPL+D) by driving a cloth simulator with a smooth, physically plausible transition between the two bodies.

\paragraph{\textbf{Source-to-target body interpolation.}}
Let the source proxy inside the garment be SMPL with parameters $(\theta_{\mathrm{c}}, \beta_{\mathrm{c}})$, and let the target proxy be SMPL+D with parameters $(\theta_{\mathrm{b}}, \beta_{\mathrm{b}}, d, s)$. Since SMPL is a special case of SMPL+D,
\begin{equation}
(\theta_{\mathrm{c}}, \beta_{\mathrm{c}})\ \equiv\ (\theta_{\mathrm{c}}, \beta_{\mathrm{c}}, d=\mathbf{0}, s=(1,1,1)),
\end{equation}
we construct a short transition sequence of $T$ frames indexed by a scalar schedule $\alpha_t \in [0,1]$ that interpolates the underlying body representation:
\begin{equation}
\beta_t = (1-\alpha_t)\beta_{\mathrm{c}} + \alpha_t \beta_{\mathrm{b}},\quad
d_t = \alpha_t d,\quad
s_t = (1-\alpha_t)\mathbf{1} + \alpha_t s .
\end{equation}
For pose, we interpolate joint rotations from $\theta_{\mathrm{c}}$ to $\theta_{\mathrm{b}}$ using quaternion slerp, producing a continuous motion that avoids abrupt changes in pose. The gradual transition improves simulation robustness by avoiding sudden body swaps that would otherwise induce heavy interpenetrations and unstable solver behavior; thus, the garment adapts progressively with the body transition.


\paragraph{\textbf{Cloth simulation with smooth body transition.}}
Given the transition body sequence $\{B_t\}_{t=1}^{T}$ produced from $(\theta_t,\beta_t,d_t,s_t)$ and the input garment mesh, we employ ContourCraft~\cite{grigorev2024contourcraft} to simulate garment deformation over the sequence. During the simulation, the network predicts per-vertex accelerations on the garment connectivity graph and explicitly models cloth-body and cloth-cloth interactions, making it suitable for multi-layer and multi-garment outfits. In practice, the model does not require a manifold surface representation as long as mesh connectivity is defined. We take the final frame $G_T$ as the dressed garment on the target SMPL+D body, and the resulting dressed state can be reused for downstream tasks.

\begin{figure*}
    \centering
    \includegraphics[width=1\linewidth]{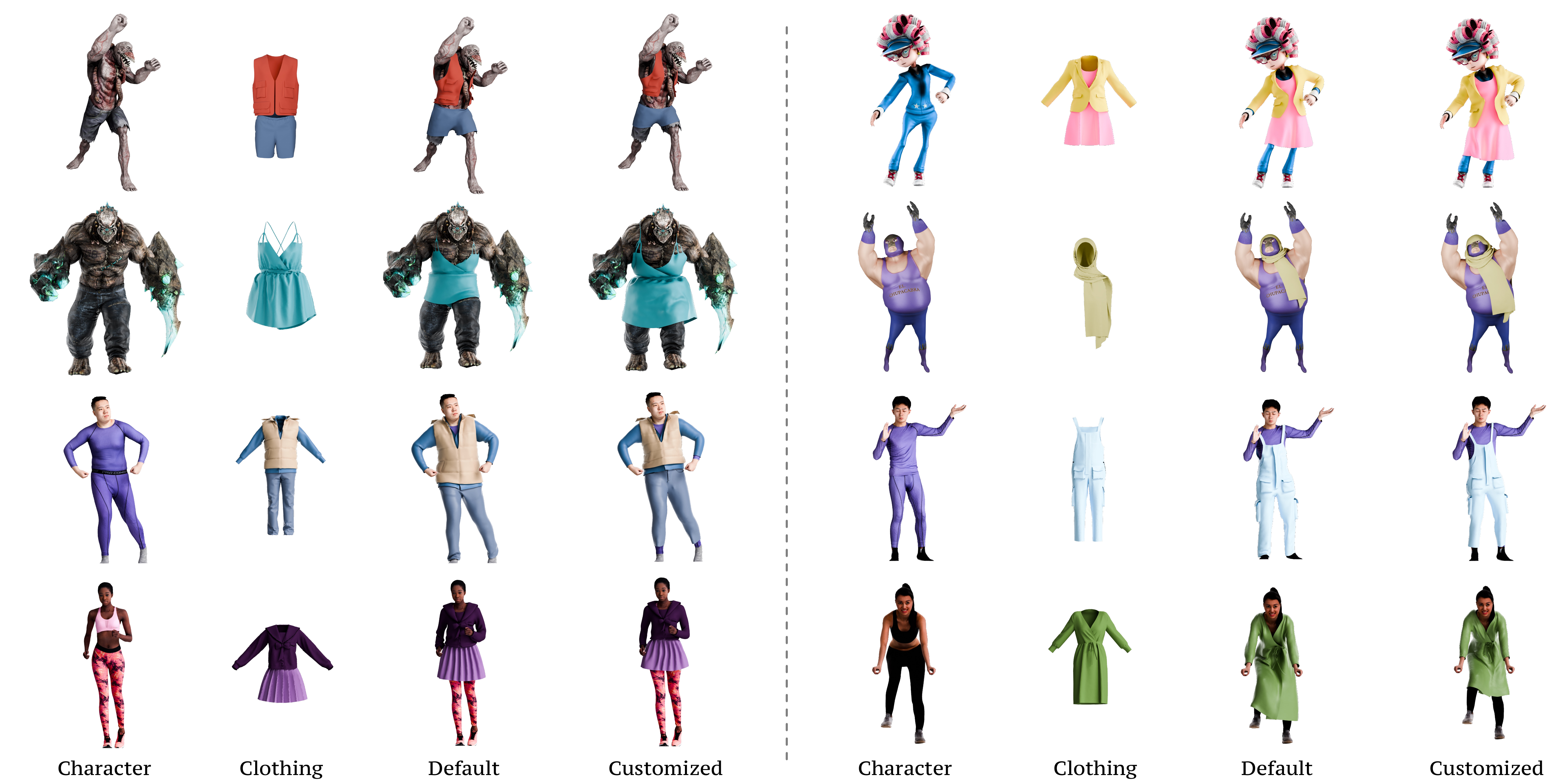}
    \caption{Our system can dress a wide range of garments on various humanoid bodies in diverse poses. The default mode preserves the original garment size, and the customized mode allows users to flexibly adjust the size. }
    \label{fig:customize}
\end{figure*}
\paragraph{\textbf{User-controlled garment size adjustment}}
We allow users to adjust the overall garment size to achieve desired try-on effects. A user-specified anisotropic scaling is applied to garment vertices, thereby scaling edge lengths, which the simulator uses as reference constraints during simulation. This modification only affects the garment rest configuration, reusing the same clothing-body correspondences and registration.

\section{Experiments}

We evaluate each stage of our pipeline, compare with state-of-the-art methods, and conduct ablations to analyze key design choices. Dataset and implementation details are in \ASection{Suppl. A} and \ASection{Suppl. B}, while expert and user study results are in \ASection{Suppl. D.5} and \ASection{Suppl. F}.

\subsection{Performance measurement}

\revised{We conducted all experiments on an NVIDIA GeForce RTX 4090 GPU with 64GB of RAM. The system achieves an approximate runtime of 4.5 minutes to dress a character. Specifically, clothing-SMPL correspondence prediction takes around 20 seconds, clothing-SMPL registration requires another 40 seconds, body correspondence prediction and body-SMPL+D registration together account for approximately 180 seconds, and clothing fitting is completed in about 15 seconds.}

\subsection{Qualitative Results}

\cref{fig:customize} demonstrates qualitative results of our system's default mode, and customized mode. It preserves garment shape in default mode, and supports user-defined scaling in customized mode. 
Our method handles diverse garment types and character shapes, including bulky, non-human, and stylized bodies, showcasing its robustness and broad applicability. Additional results are provided in \ASection{Suppl. C}.

\subsection{Comparisons with Existing Approaches}
For comparative evaluation on overall clothing draping performance, we evaluate our method on the Cloth3D~\cite{bertiche2020cloth3d} test set against DrapeNet~\cite{de2023drapenet} and ISP~\cite{li2024isp}, using 100 top-trouser pairs across 4 diverse poses each. Metrics include Chamfer Distance (CD), Point-to-Mesh (P2M), and Interpenetration Ratio (IR). Since the two methods can only take parametric bodies (SMPL) as input, for a fair comparison, we omit the \textit{Body-SMPL correspondence prediction} and \textit{Body-SMPL+D registration} stages from our pipeline and also use SMPL parameters as input.

\begin{table}[h]
\centering
\caption{Comparison of clothing draping performance.}
\resizebox{\linewidth}{!}{%
\begin{tabular}{lccccccc}
\toprule
\multirow{2}{*}{\textbf{Method}} & \multicolumn{3}{c}{\textbf{Top}} & & \multicolumn{3}{c}{\textbf{Trousers}} \\
\cmidrule(lr){2-4} \cmidrule(lr){6-8}
 & CD$\downarrow$ & P2M$\downarrow$ & IR(\%)$\downarrow$ & & CD$\downarrow$ & P2M$\downarrow$ & IR(\%)$\downarrow$ \\
\midrule
DrapeNet & 0.0101 & 0.0100 & 1.58 & & 0.0013 & 0.0012 & 4.12 \\
\revised{DrapeNet (post)} & \revised{0.0270} & \revised{0.0267} & \revised{\textbf{0.43}} & & \revised{0.0022} & \revised{0.0020} & \revised{0.66} \\
ISP & 0.0034 & 0.0032 & 13.02 & & 0.0081 & 0.0080 & 22.20 \\
\revised{ISP (post)} & \revised{0.0057} & \revised{0.0054} & \revised{12.03} & & \revised{0.0098} & \revised{0.0096} & \revised{13.74} \\
Ours & \textbf{0.0008} & \textbf{0.0007} & 0.56& & \textbf{0.0006} & \textbf{0.0005} & \textbf{0.20} \\
\bottomrule
\end{tabular}%
}
\label{tab:comparison_results}
\end{table}

\begin{figure}[H]
    \centering
    \includegraphics[width=1\linewidth]{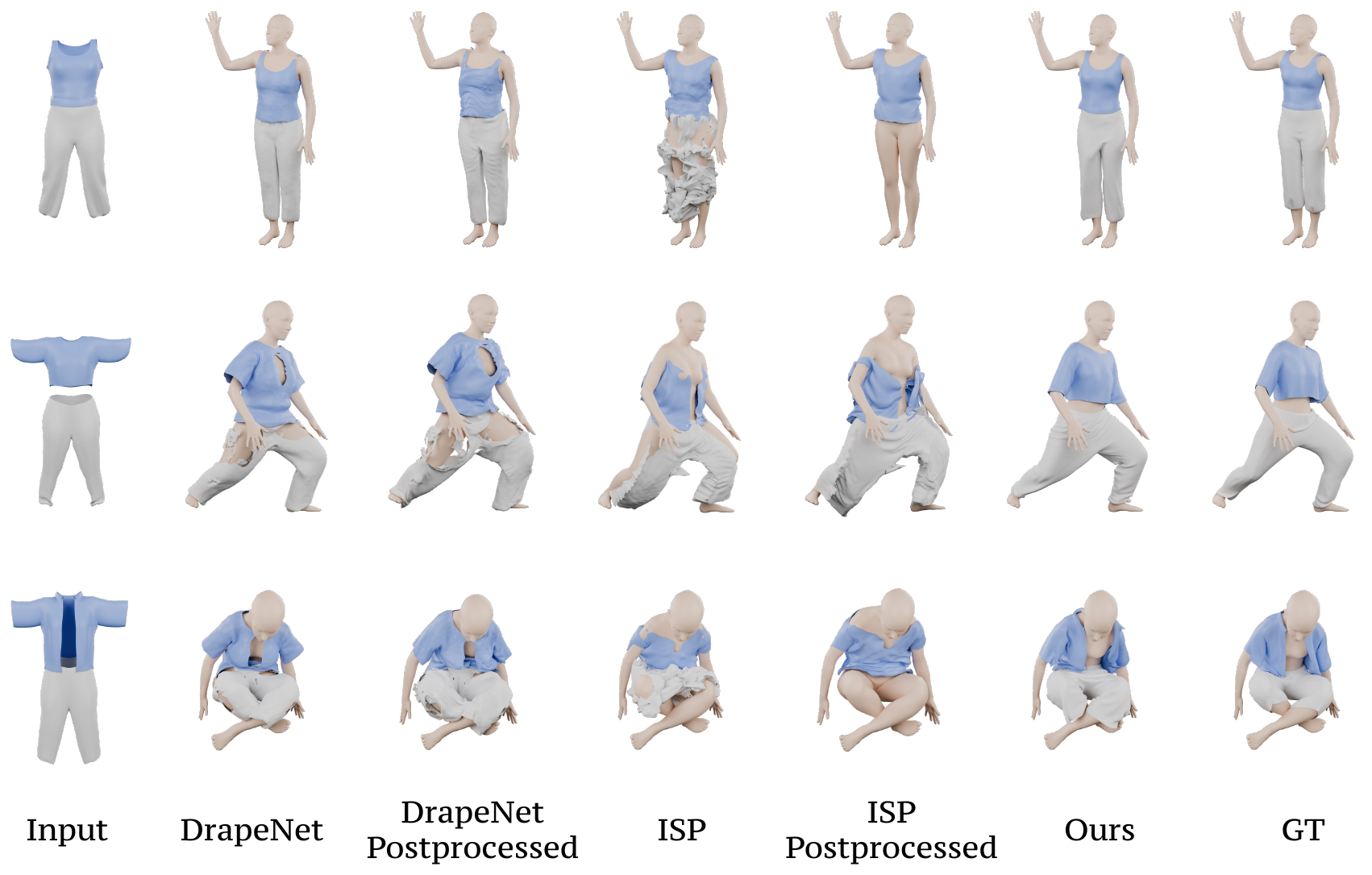}
    \caption{\revised{Qualitative comparisons with DrapeNet and ISP.}}
    \label{fig:DrapeNet}
\end{figure}

As shown in \cref{tab:comparison_results}, our method achieves the lowest CD, P2M, and IR. Visual results in \cref{fig:DrapeNet} further confirm that our predictions closely match the ground truth and exhibit the fewest artifacts among all methods. Additional results are provided in \ASection{Suppl. D.1}. 

\revised{For a fair comparison, we also applied postprocessing to the results of DrapeNet and ISP by introducing additional simulation steps to alleviate overall geometric distortions. While this leads to a noticeable reduction in IR and visually mitigates interpenetration, it does not recover failed try-on cases. Both DrapeNet and ISP output garments with different topology than the input because they reconstruct the garment surface from an implicit representation rather than deforming the original mesh, so no vertex connectivity or topology is preserved. As a result, the input garment mesh cannot be used as a reference during postprocessing simulation, limiting the effectiveness of subsequent correction. In addition, if the results are too poor, the simulation will fail, such as the pants in the first and third examples of ISP in \cref{fig:DrapeNet}. }

\begin{figure}[htbp]
    \centering
    \includegraphics[width=0.82\linewidth]{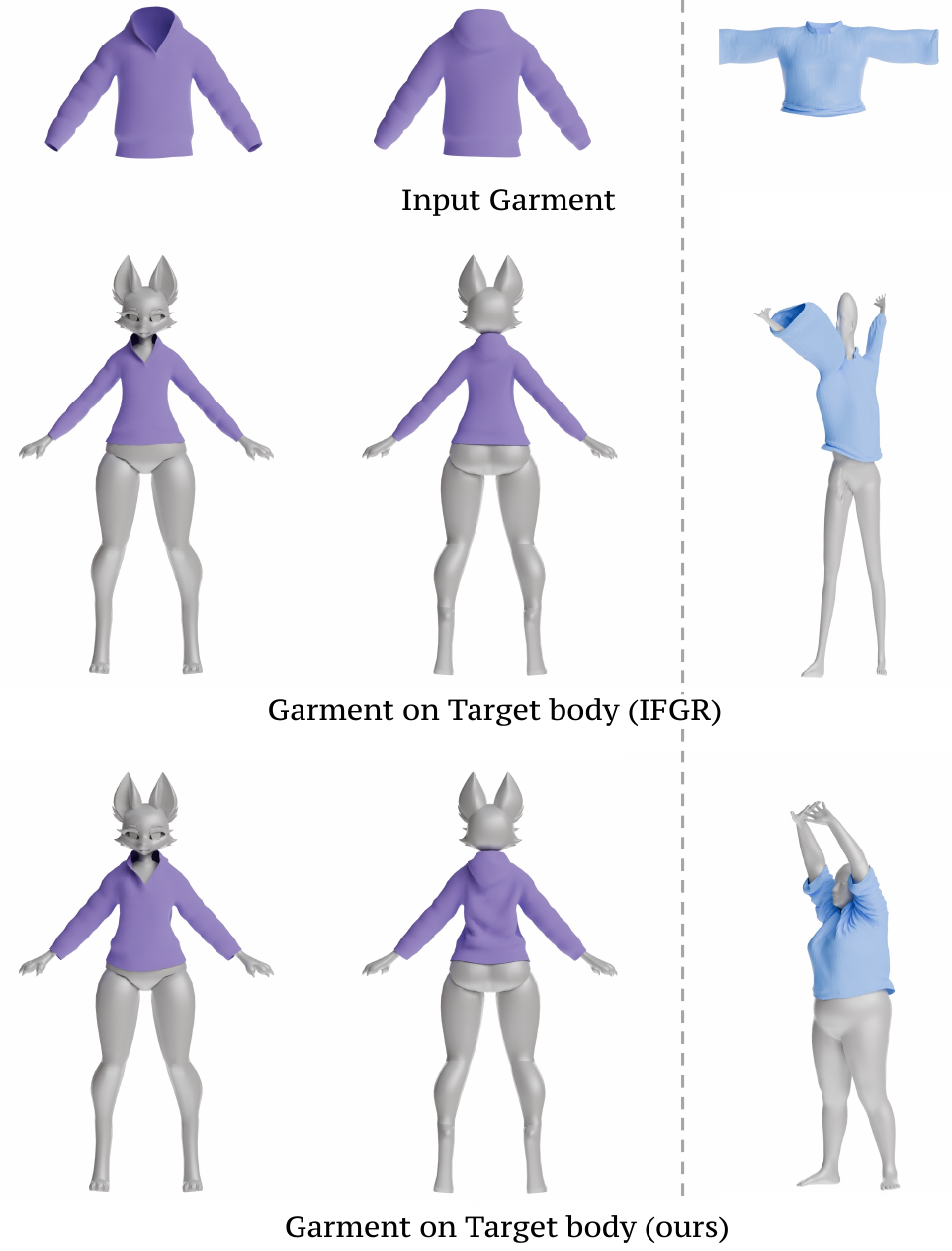}
    \caption{Comparison to IFGR~\cite{Intersection-Free_Garment_Retargeting}. Left: pose-matched retargeting. Right: large pose discrepancy; IFGR converges to an intermediate inflated avatar under collision constraints.}
    \label{fig:IFGR}
\end{figure}

We further compare our method with an optimization-based retargeting method, Intersection-Free Garment Retargeting (IFGR) ~\cite{Intersection-Free_Garment_Retargeting}, which transfers garments with skeletons to rigged characters. IFGR assumes a manifold input garment and relies on pose compatibility between source and target to ensure a stable retargeting~\cite{Intersection-Free_Garment_Retargeting}. 
\rl{As shown in \cref{fig:IFGR} (left), under pose-matched settings, both methods produce plausible transfers, while our results additionally exhibit simulator-driven, physically plausible draping details such as wrinkles. However, when applied to cases with large pose discrepancies (e.g., a T-pose garment transferred to a lifting-pose target as shown in \cref{fig:IFGR} right), which violate IFGR’s pose-compatibility assumption, IFGR fails to produce correct transfers. IFGR inflates a skeleton-initialized proxy toward the posed target via collision-constrained optimization. Under large pose gaps, the optimization converges to an intermediate body configuration instead of the target pose due to high garment–body contact penalties. In contrast, our method transfers garments accurately through simulation along a smooth SMPL shape-and-pose transition, providing a well-conditioned motion sequence and avoiding abrupt body changes that typically lead to severe interpenetrations.}


\begin{figure*}[htbp]
    \centering
    \includegraphics[width=1\linewidth]{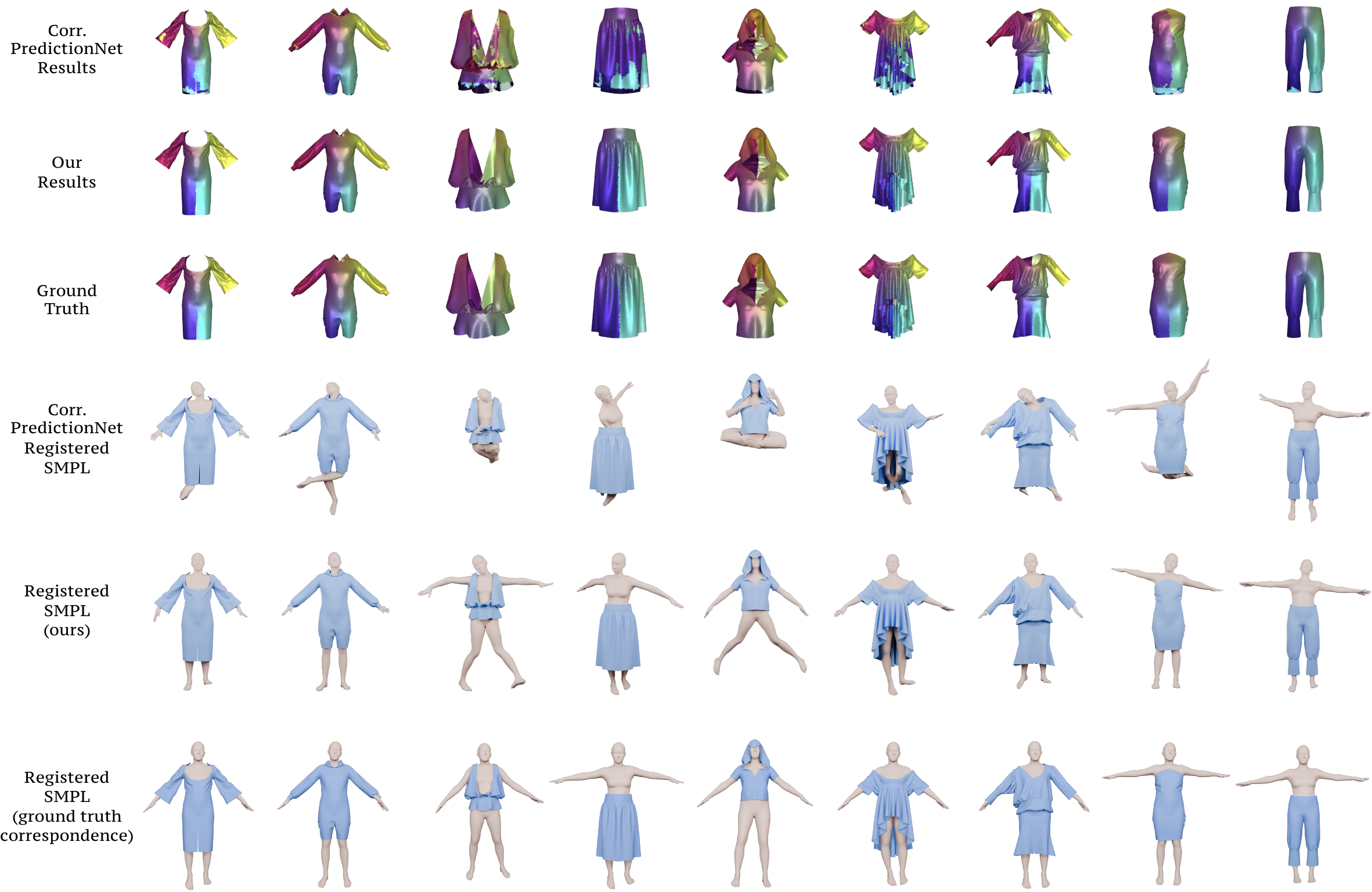}
    \caption{Qualitative comparison of clothing-SMPL correspondence. Compared to using CorrPredNet (top), our method (middle) yields more accurate correspondence results and SMPL registrations. \revised{The bottom row shows the registration results using ground-truth correspondence.}}
    \label{fig:clothing_corres_ablation}
\end{figure*}

\begin{figure}[!ht]
    \centering
    \includegraphics[width=1\linewidth]{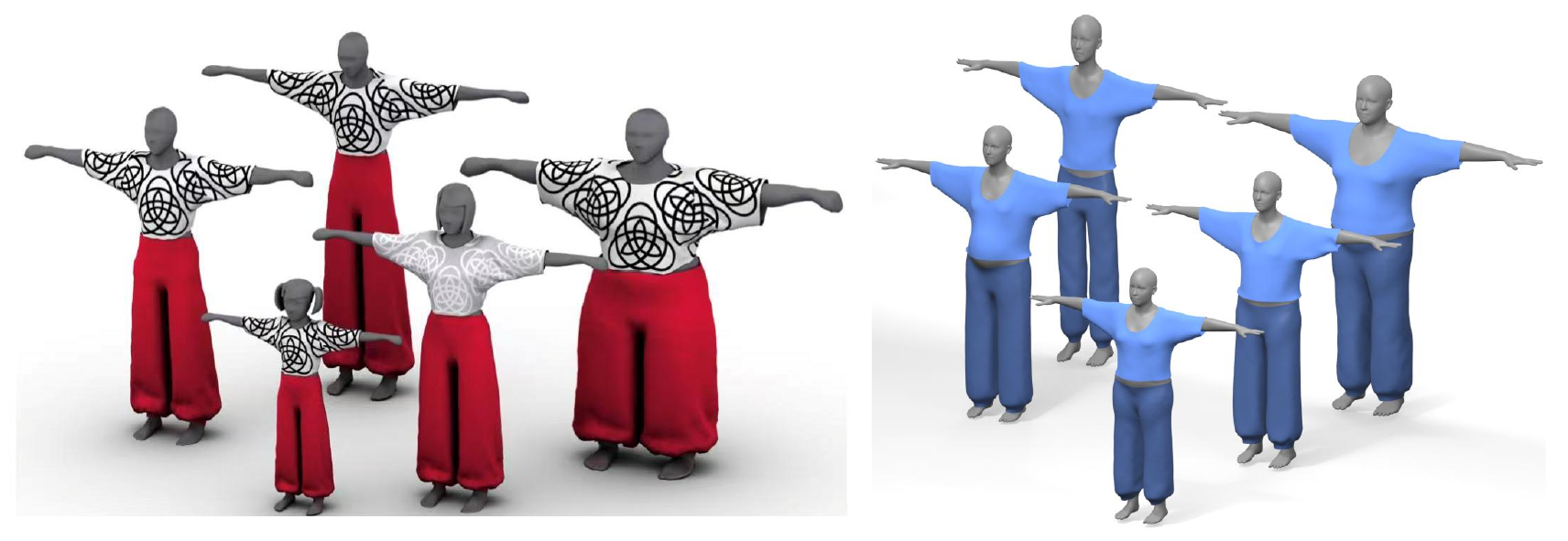}
    \caption{\revised{Qualitative comparison between DPGT (left) and our method (right).}}
    \label{fig:comparison_DPGT}
\end{figure}
\revised{We provide a visual comparison with Design Preserving Garment Transfer (DPGT)~\cite{brouet2012design} using structurally similar garments and poses. DPGT assumes reliable source–target body correspondence and compatible rigs as extra inputs, and can only apply to bodies (source and target) with similar poses, whereas our method addresses a more general setting where such assumptions may not hold; when a source mannequin is available, we can still handle it as a special case, as shown in~\cref{fig:comparison_DPGT}.}

\subsection{Evaluation on Clothing-SMPL Correspondence and Registration}\label{sec:eva_clothing}

We evaluate our clothing-SMPL correspondence method on 100 garment-body pairs from the GarmentCode dataset~\cite{GarmentCodeData:2024}, where each sample consists of a garment mesh and its paired SMPL body. Performance is evaluated using Mean Euclidean Error (MEE) and Mean Geodesic Error (MGE) between predicted and ground-truth correspondences, and Interpenetration Ratio (IR) after registration to quantify physical realism.


\begin{table}[htbp]
\centering
\small
\caption{Comparison of clothing-SMPL correspondence and registration.}
\label{tab:cloth_correspondence}
\begin{tabular}{lcccc}
\toprule
\textbf{Method} & \textbf{MGE $\downarrow$} & \textbf{MEE $\downarrow$} & \textbf{IR (\%) $\downarrow$} & \textbf{No-Penetration Rate $\uparrow$} \\
\midrule
CorrPredNet & 0.2290 & 0.0993 & 1.02 & 5\% \\
Ours & \textbf{0.0499} & \textbf{0.0274} & \textbf{0.17} & \textbf{59\%} \\
\midrule
\revised{GT Corr.} & \revised{0} & \revised{0} & \revised{0.11} & \revised{65\%} \\
\bottomrule
\end{tabular}
\end{table}

We adopt the training-free CorrPredNet (originally developed for body correspondence) as a baseline since it provides an alternative possibility for unifying the framework. As shown in \cref{tab:cloth_correspondence}, our method significantly outperforms CorrPredNet. Our method yields zero interpenetration in 59\% of cases, while CorrPredNet struggles with partial-to-complete mappings, resulting in inaccurate correspondences and unrealistic SMPL fits.

Qualitative results in \cref{fig:clothing_corres_ablation} further illustrate this difference. For instance, in the fourth column, CorrPredNet incorrectly maps the lower hem of the dress to the feet, distorting the SMPL pose. Our method avoids such errors and produces clean, well-registered fits across diverse garment types. 

\revised{We also use the ground truth correspondence for the clothing-SMPL registration. Comparisons in \cref{tab:cloth_correspondence} and \cref{fig:clothing_corres_ablation} indicate that our predicted correspondences approach the quality of the ground truth, yielding competitive Intersection Ratio (IR), No-Penetration Rate, and visual quality in subsequent Clothing-SMPL registration.}

\subsection{Evaluation on Body-SMPL Correspondence and Registration}\label{sec:eva_body}

We introduce a new benchmark of eight stylized humanoid characters for a challenging evaluation, each with ten randomly sampled diverse poses. Details are provided in \ASection{Suppl. A.2}


\paragraph{\textbf{Body-SMPL Correspondence Evaluation.}}
We evaluate our meth-od against GeomFmaps~\cite{donati2020deepGeoMaps}, 
ULRSSM~\cite{cao2023unsupervised}, HybridFmaps~\cite{bastianxie2024hybrid}, DiffusionNet~\cite{sharp2022diffusionnet}, and Diff3f~\cite{dutt2024diffusion}. 

\begin{figure}[ht]
    \centering
    \includegraphics[width=0.9\linewidth]{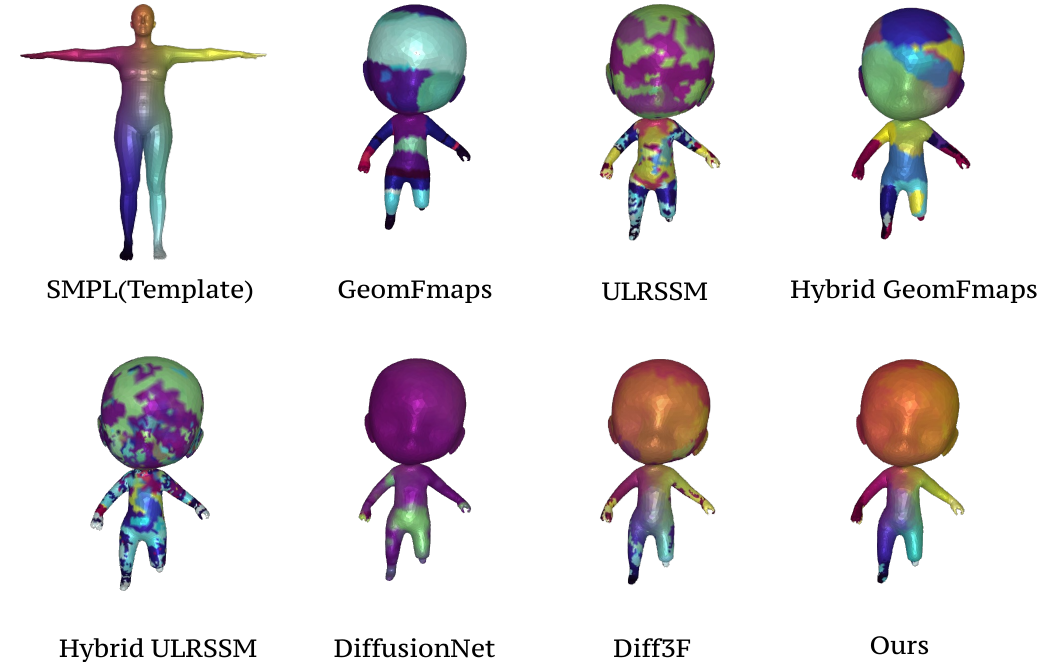}
    \caption{Qualitative comparison of GeomFmaps, ULRSSM, Hybrid GeomFmaps (Hybrid GFM), Hybrid ULRSSM, DiffusionNet, Diff3f, and our method predicting body correspondence.}
    \label{body_correspondence_main}
\end{figure}

DiffusionNet is trained on 2,000 samples using our supervision strategy (\cref{sec:clothing_corr}), while others use public checkpoints. As shown in \cref{tab:mean_body_geodesic_errors_main}, our method achieves the lowest average MEE and MGE. Furthermore, functional map methods struggle with unseen topologies; DiffusionNet lacks robustness to shape and pose, and Diff3f improves with semantic priors. Our method further improves coherence and consistency (see \cref{body_correspondence_main}).


\begin{table}[h]
\centering
\small
\caption{Comparison of body-SMPL correspondence performance.}
\begin{tabular}{lcc}
\toprule
\textbf{Method} & \textbf{MEE($\times$ 10) $\downarrow$} & \textbf{MGE($\times$ 10) $\downarrow$} \\
\midrule
GeomFmaps         & 3.400 & 6.867 \\
ULRSSM            & 3.224 & 6.442 \\
Hybrid GeomFmaps  & 3.429 & 6.966 \\
Hybrid ULRSSM     & 3.215 & 6.549 \\
DiffusionNet      & 5.270 & 5.900 \\
Diff3f            & 2.320 & 3.110 \\
Ours     & \textbf{1.700} & \textbf{2.100} \\
\bottomrule
\end{tabular}
\label{tab:mean_body_geodesic_errors_main}
\end{table}


\paragraph{\textbf{Body-SMPL Registration Evaluation}}

\begin{figure}[ht]
    \centering
    \includegraphics[width=0.99\linewidth]{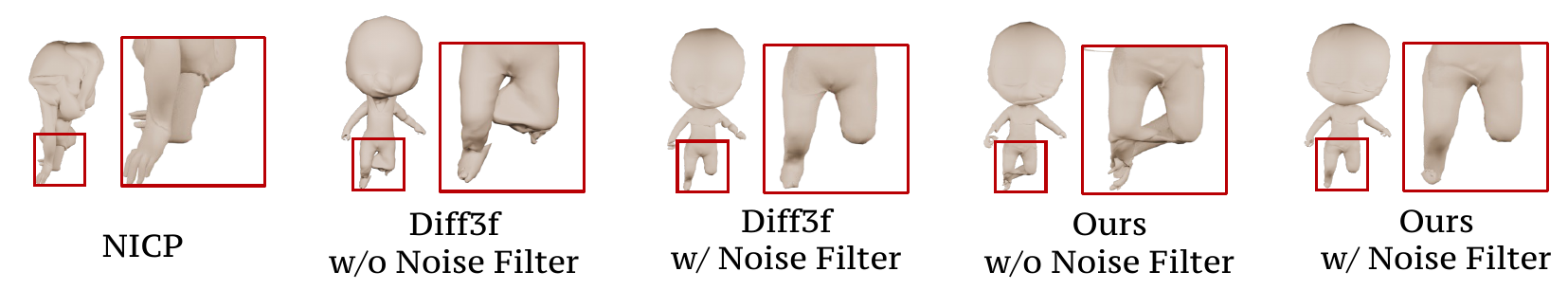}
    \caption{Qualitative comparisons of the performance of Body-SMPL+D registration.}
    \label{fig:body-SMPL-Registration-main}
\end{figure}

We evaluate SMPL registration accuracy against NICP~\cite{marin2025nicp} and Diff3f~\cite{dutt2024diffusion} using Chamfer Distance between the registered SMPL+D and input body mesh. 
As shown in \cref{tab:sota_chamfer_distance}, our approach achieves the lowest CD, indicating superior registration quality. Furthermore, the proposed noise filtering module plays a critical role in enhancing alignment accuracy. Visual comparisons in \cref{fig:body-SMPL-Registration-main} show that NICP fails to generalize to the character with extreme body shapes (big head and small body), while Diff3f exhibits limitations under large pose deformations. In contrast, our method demonstrates consistently robust and precise registrations across diverse test cases. Notably, even without noise filtering, our method outperforms all baselines, and the integration of the filter further boosts performance. More discussions can be found in \ASection{Suppl. D.3}.

\begin{table}[H]
\centering
\small
\caption{Comparison of Chamfer Distance for body-SMPL registration.}
\resizebox{\linewidth}{!}{%
\begin{tabular}{lccccc}
\toprule
\textbf{Method} & \textbf{NICP} & \makecell{\textbf{Diff3f} \\ \textbf{(w/o Filter)}} & \makecell{\textbf{Diff3f} \\ \textbf{(w/ Filter)}} & \makecell{\textbf{Ours} \\ \textbf{(w/o Filter)}} & \makecell{\textbf{Ours} \\ \textbf{(w/ Filter)}} \\
\midrule
\textbf{CD$\times 10^{-4} \downarrow$} & 13.05 & 4.99 & 4.49 & 1.46 & \textbf{1.19} \\
\bottomrule
\end{tabular}
\label{tab:sota_chamfer_distance}
}
\end{table}

\begin{figure*}
    \centering
    \includegraphics[width=0.95\linewidth]{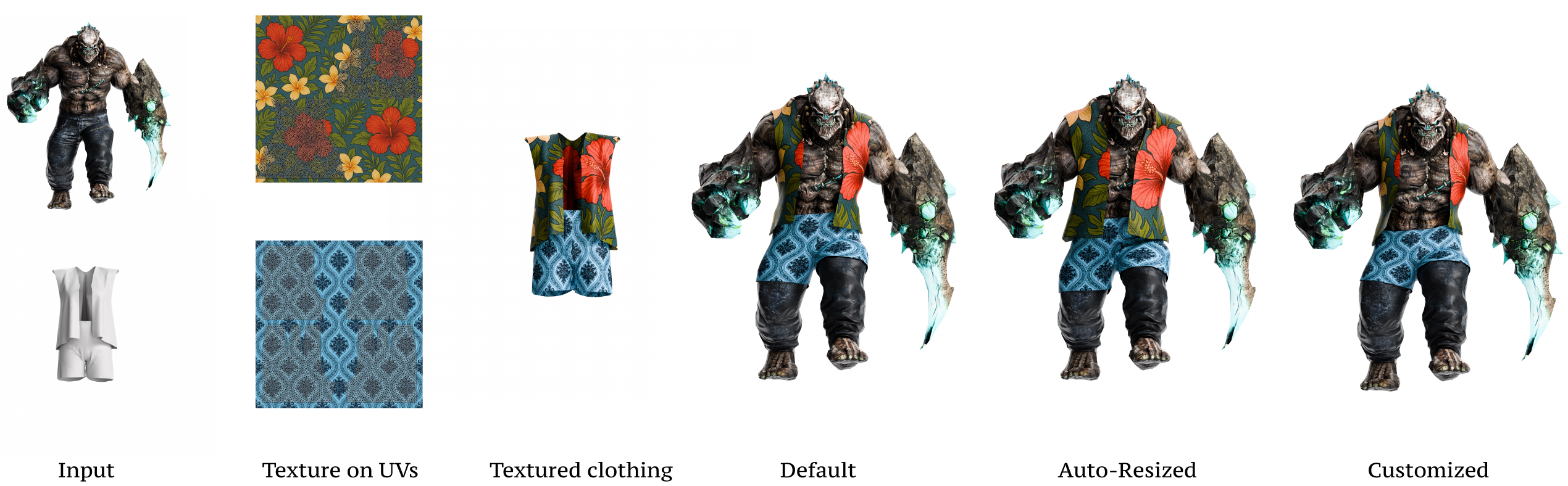}
    \caption{Draped garment with texture}
    \label{fig:texture}
\end{figure*}
\subsection{Ablation Study on Correspondence Modules}
\begin{figure}[ht]
    \centering
    \includegraphics[width=0.9\linewidth]{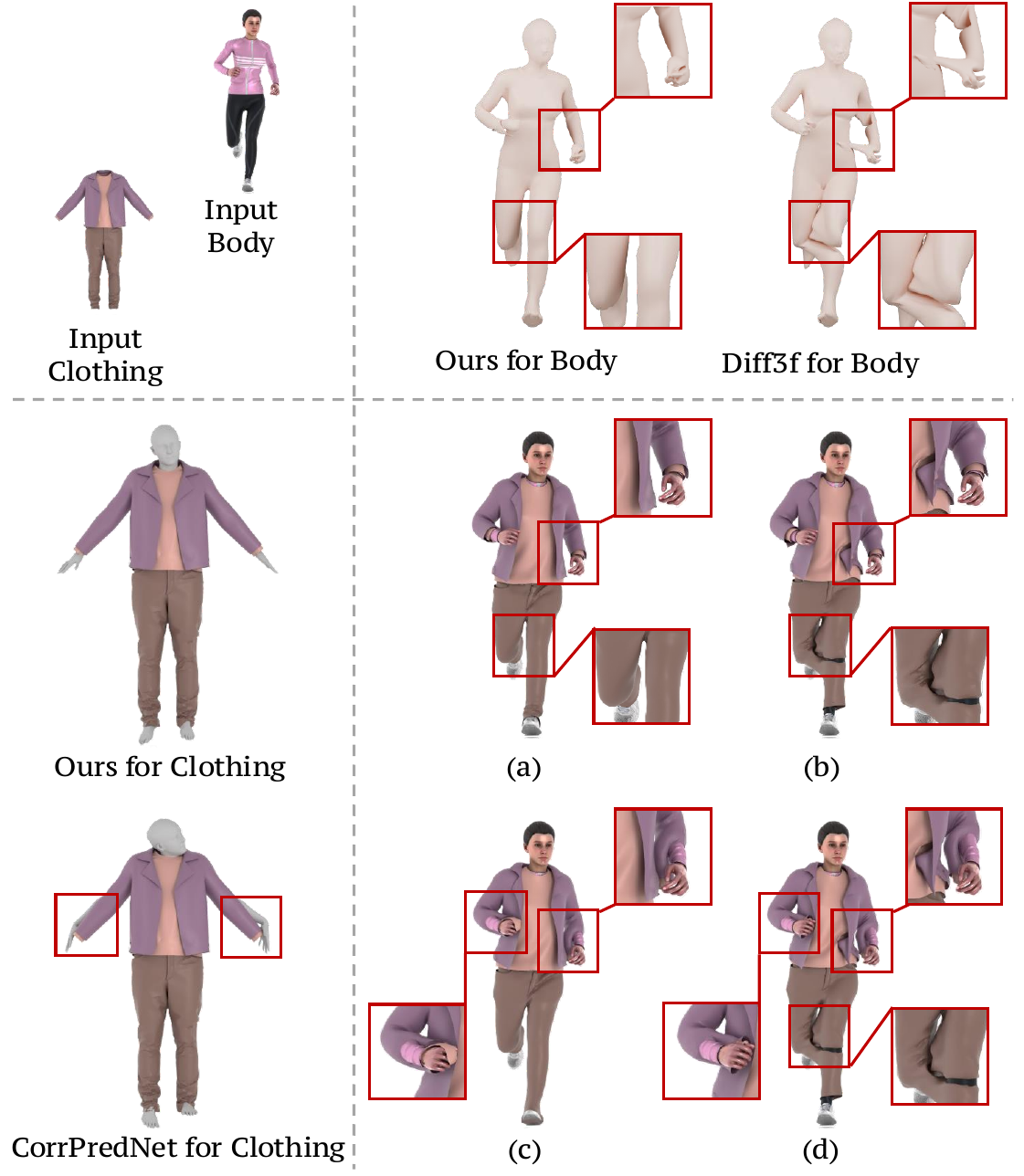}
    \caption{Ablation on Correspondence Module.}
    \label{fig:ablation_system}
\end{figure}

We assess the impact of correspondence quality on final garment fitting by replacing our body-to-SMPL and clothing-to-SMPL modules with next-best alternatives, Diff3f and CorrPredNet, respectively. This ablation focuses on the effect of each correspondence module on the final visual outcome. As shown in \cref{fig:ablation_system}, inaccurate body correspondence (b) leads to incorrect SMPL shape and poor clothing deformation, while inaccurate clothing correspondence (c) causes severe interpenetration, such as limbs poking through the garment. When both are suboptimal (d), the errors compound, resulting in the worst overall fitting. In contrast, our full method (a) achieves realistic draping. Additional results are provided in \ASection{Suppl. E.5, F.5}.

\subsection{Applications}
\begin{figure*}[htbp]
    \centering
    \includegraphics[width=0.92\linewidth]{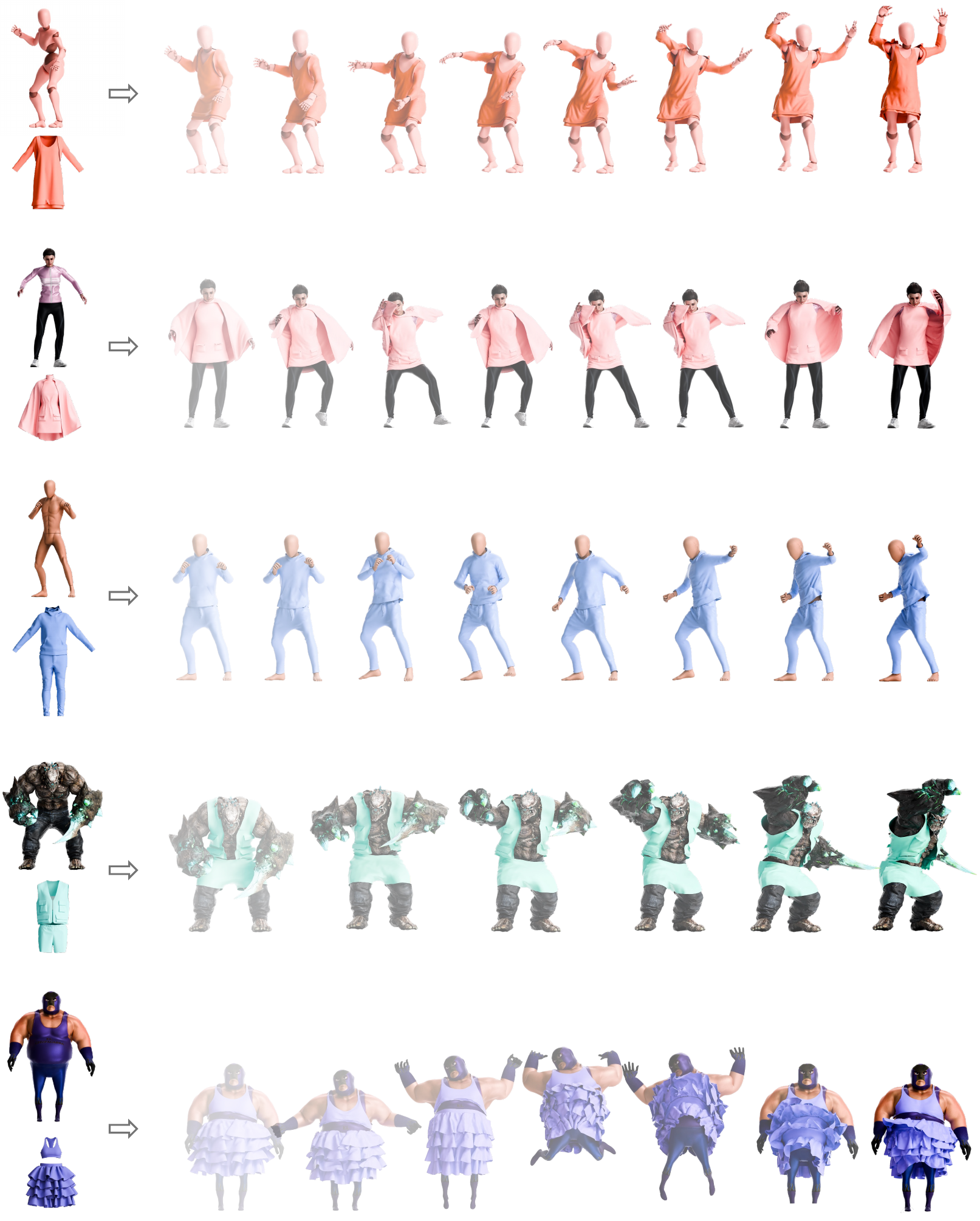}
    \caption{Our system can be applied to downstream tasks like dressing up a character for animation.}
    \label{fig:animation_main}
\end{figure*}

Our system enables garments to be automatically draped onto posed characters, making it well-suited for animation initialization and digital human content creation. By removing the need for manual garment initialization, it simplifies character creation pipelines and supports rapid prototyping of clothed digital humans for films, games, and virtual reality applications. As shown in \cref{fig:animation_main}, garments can be fitted to the first frame of a motion sequence and then simulated in Marvelous Designer~\cite{marvelousdesigner}. This also makes the pipeline more accessible to non-expert users while providing professional artists with efficient tools for complex character development.

\begin{figure}[htbp]
    \centering
    \includegraphics[width=0.9\linewidth]{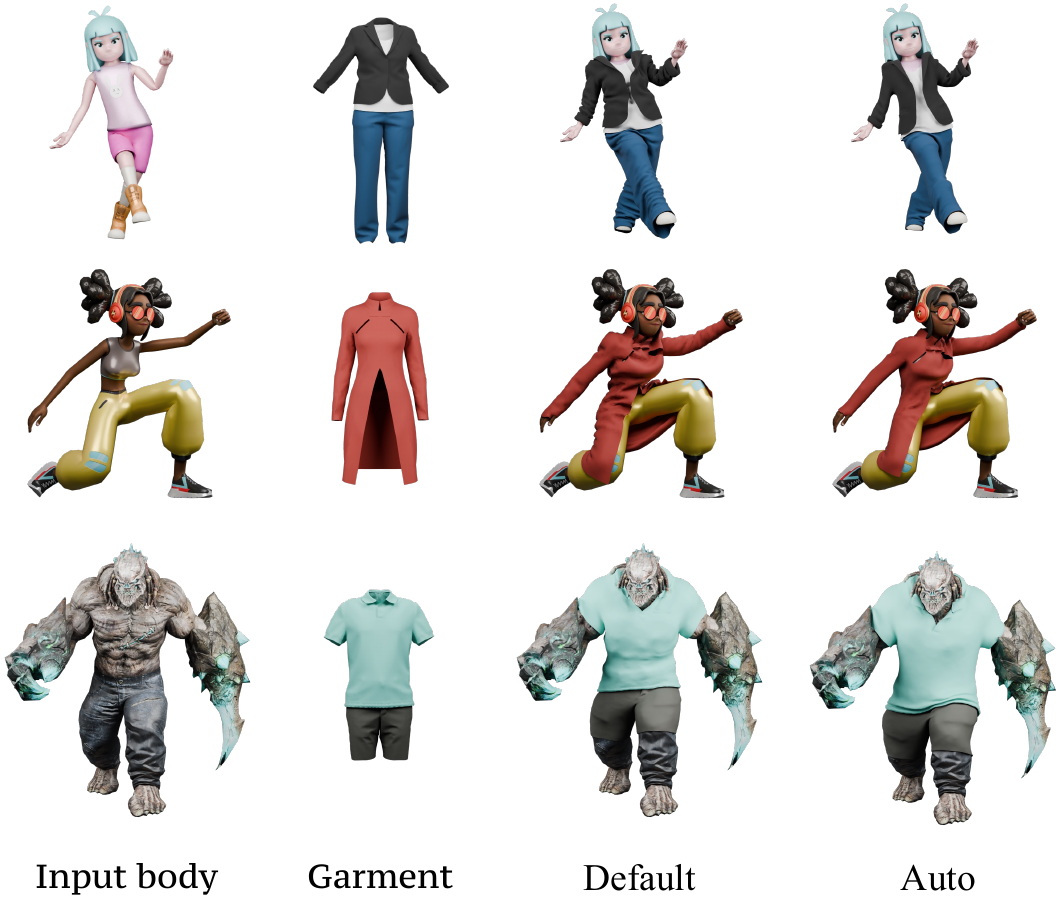}
    \caption{Comparisons between default mode and auto-resize mode}
    \label{fig:auto_resize}
\end{figure}

Our system also supports automatic garment size adjustment using the anisotropic body scale $s$ estimated during body-SMPL+D registration. Specifically, we scale the garment rest configuration along the $x$, $y$, and $z$ axes according to $s$, which correspondingly scales the rest geometry and the rest edge lengths used as reference constraints during simulation. This provides a lightweight mechanism to adapt garment size to different body proportions without recomputing correspondences or rerunning registration. \cref{fig:auto_resize} compares the two modes: \emph{default} (original rest shape), \emph{auto-resize} (scaled by $s$).

Our method further supports large-scale generation of synthetic clothed human scans (\cref{fig:customize}). Starting from a small set of tightly clothed scans, we can automatically drape garments onto the scans to create diverse clothed identities. Importantly, our system preserves the original garment design and UV parameterization, so textures authored in UV space are faithfully retained on the final garments, as shown in \cref{fig:texture}. This could potentially allow the generation of textured clothed humans for downstream applications such as 3D reconstruction, pose estimation, and virtual try-on.

\begin{figure}[htbp]
    \centering
    \includegraphics[width=0.95\linewidth]{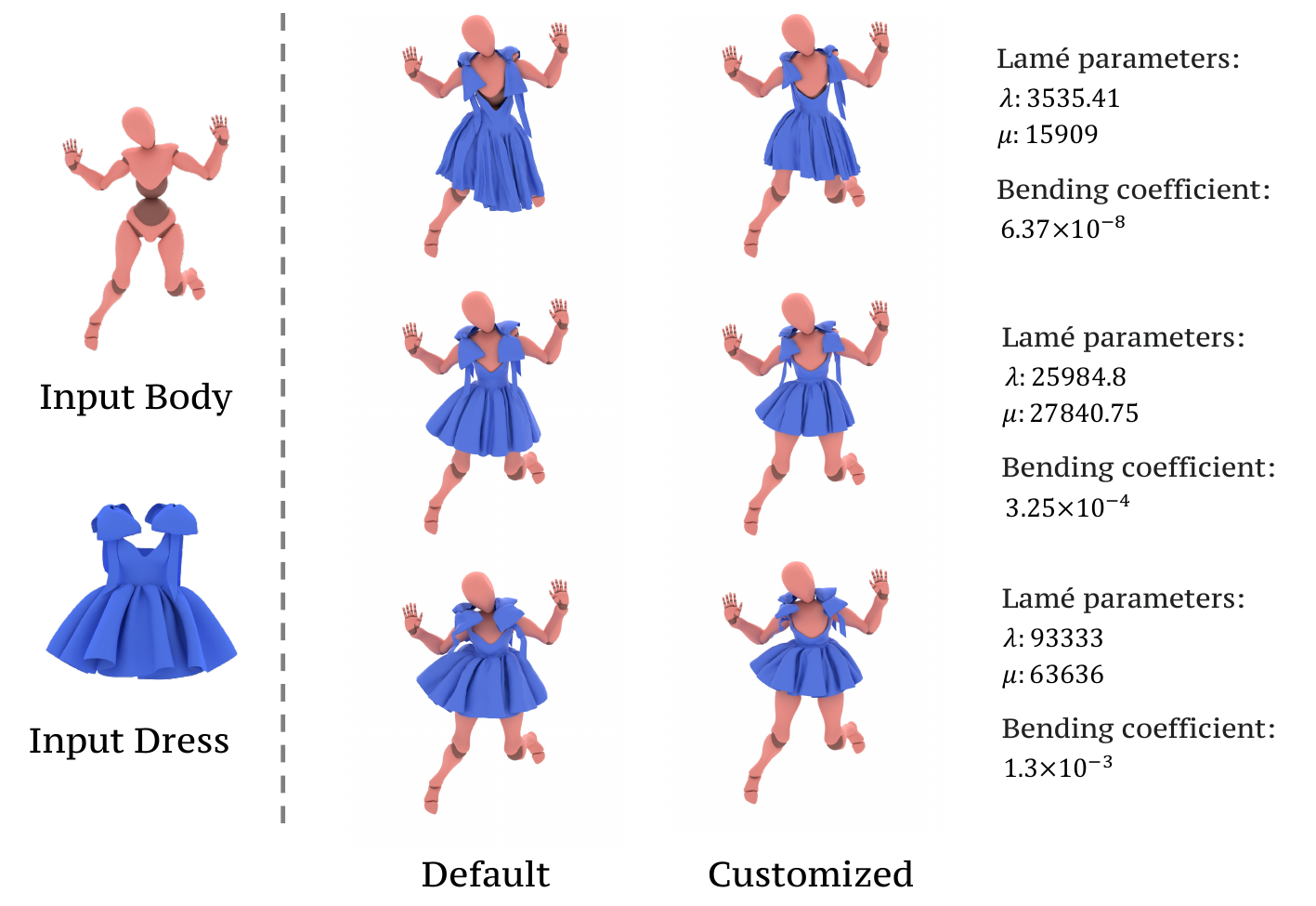}
    \caption{Effect of material parameters on a dress. As Lamé parameters $(\lambda, \mu)$ and the bending coefficient increase from top to bottom, the dress becomes stiffer, retaining more of its shape.}
    \label{fig:material_behavior}
\end{figure}

Finally, our system supports customizable material parameters for simulating different fabric types. As shown in \cref{fig:material_behavior}, increasing stiffness leads to less deformation and stronger shape retention, enabling flexible control of garment behavior under different virtual try-on scenarios.

\subsection{Limitations and Future Work}

\begin{figure}[htbp]
    \includegraphics[width=1\linewidth]{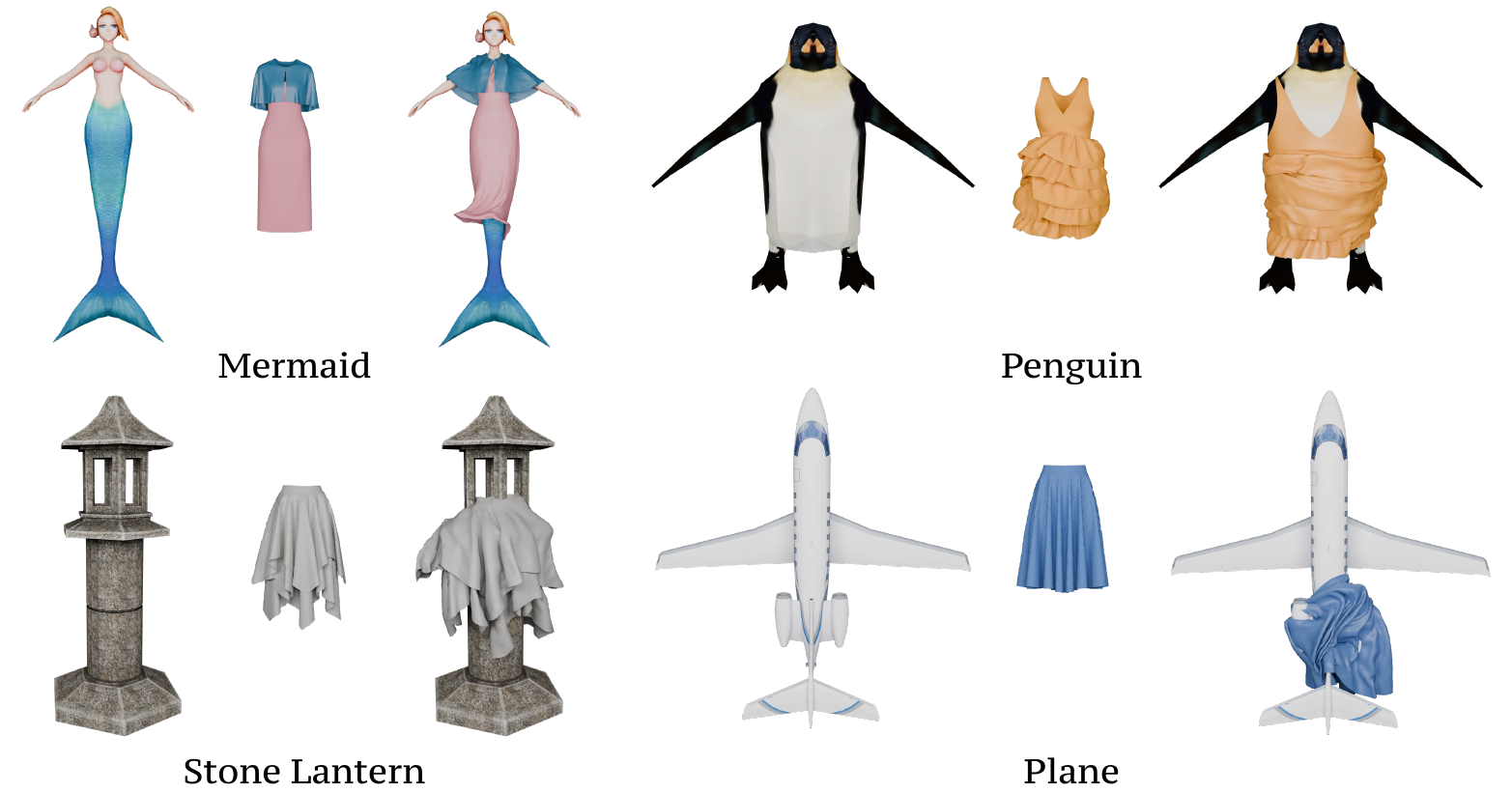}
    \caption{Applying our system to dress less humanoid models.}
    \label{fig:non_humanoid}
\end{figure}

Our system has several limitations. First, the method relies on establishing meaningful correspondences via the SMPL proxy, which restricts its applicability to shapes that are sufficiently humanoid. While models with coarse humanoid structure (e.g., mermaids or penguins) can still yield reasonable results, highly non-humanoid shapes (e.g., stone lanterns or airplanes) lead to disorganized SMPL registration and failure cases, as shown in \cref{fig:non_humanoid}. This limits the applicability of our system to assets that deviate significantly from human anatomy.

\begin{figure}[htbp]
    \includegraphics[width=1\linewidth]{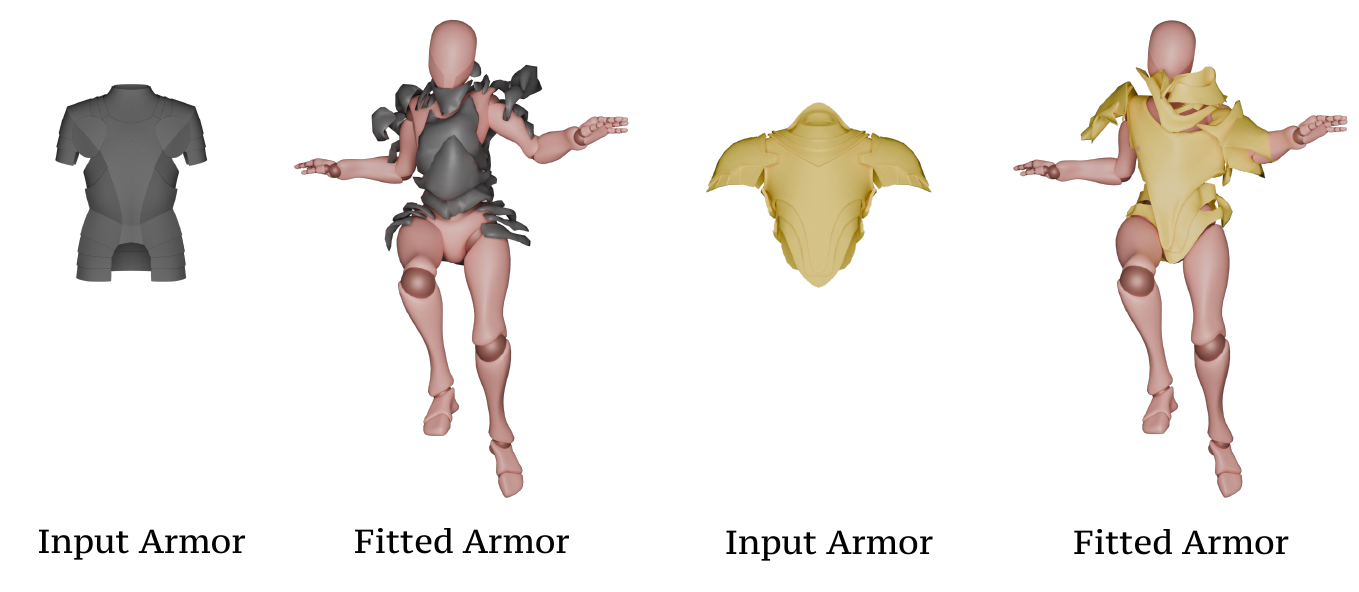}
    \caption{
    Our system fails to accurately fit hard and segmented material armors to the input body.
    }
    \label{fig:failurecases}
\end{figure}

Second, the simulation stage inherits limitations from the underlying neural cloth simulator. In particular, it struggles with hard materials (e.g., armor) and disconnected garment components (e.g., buttons, zippers), as illustrated in \cref{fig:failurecases}. Such elements are either not faithfully represented (due to soft-material assumptions) or may drift apart during simulation when not topologically connected. 

\begin{figure}[H]
    \centering
    \includegraphics[width=0.98\linewidth]{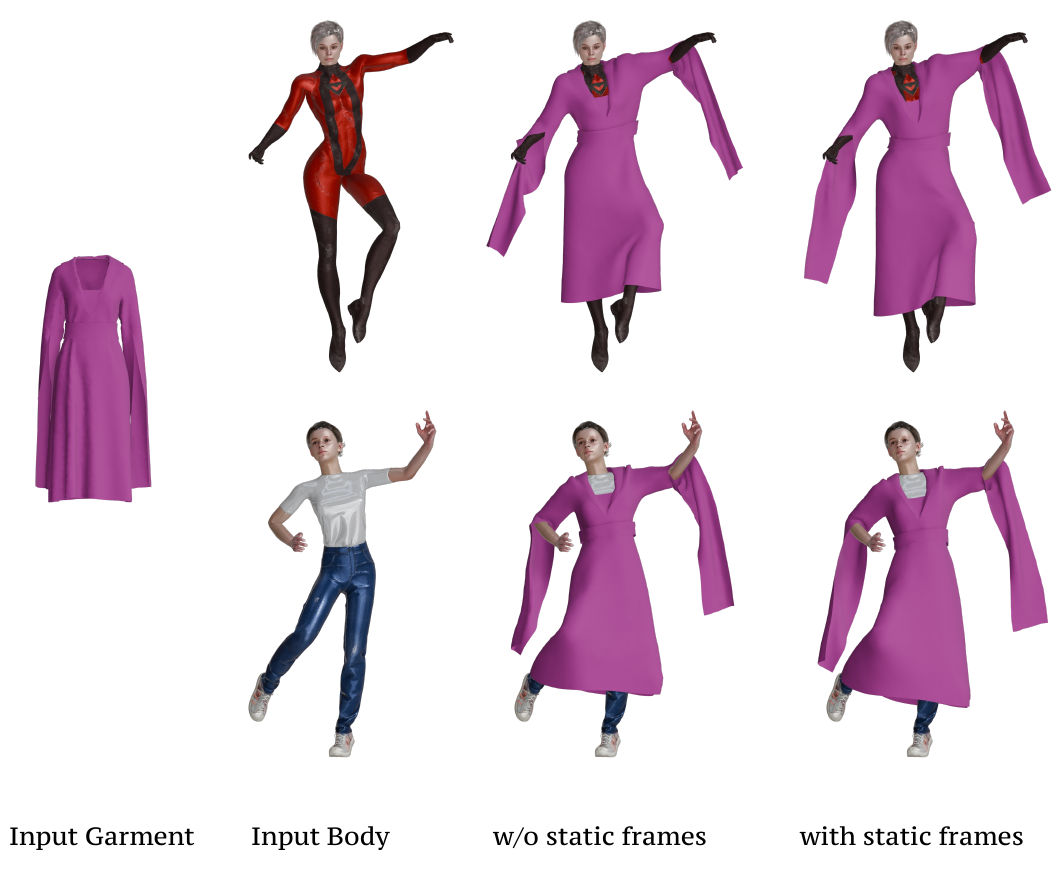}
    \caption{\revised{Examples for adding more static frames (with static frames) for simulation on our final try-on results (w/o static frames).}}
    \label{fig:static}
\end{figure}

\revised{Additionally, long and highly draping garments may not fully reach static equilibrium under the current fixed transition schedule, resulting in slight residual motion in the final frame; this can be mitigated by appending additional static simulation steps. As shown in \cref{fig:static}, adding ten additional static frames stabilizes the sleeves, improves the final try-on quality.}

Third, the current pipeline only supports clothing in canonical rest poses (A/T-pose), as the clothing-to-SMPL correspondence module is trained exclusively on such data. Extending the method to handle arbitrarily posed garments would require additional training data, which could potentially be generated using our framework in future work.

\revised{
From a physical modeling perspective, anisotropic scaling used in our auto and customized mode is not suitable for cut-and-sew manufacturing workflows; it is intended for virtual try-on scenarios where visual fit is prioritized. For workflows requiring physical realizability, users may choose uniform scaling, which preserves global proportions. Currently, the simulation directly treats input garments as rest shapes rather than relying on flat sewing patterns, which may introduce non-physical artifacts. When sewing pattern information is available, incorporating it as rest-shape constraints would improve physical realism and is a promising direction for future work.

During the simulation, two types of interpenetration may arise: (i) body–garment and (ii) body self-intersection. For (i), garment simulation is collision-aware and is performed progressively along the interpolated motion, preventing penetration in intermediate frames. For (ii), we interpolate smoothly between two SMPL poses (typically, the source SMPL is in canonical A/T pose, since the garment is in this pose). In our experiments, this gradual interpolation from a canonical-like pose did not produce observable body self-intersections. In rare scenarios where extreme poses could cause self-contact, the interpolation path may be regularized using a pose prior or collision-aware constraint.




Our pipeline is currently tailored to a collision-tolerant neural simulator (ContourCraft \cite{grigorev2024contourcraft}) and is not directly compatible with strict physics solvers such as IPC-based methods, which require intersection-free initialization. Although the observed interpenetrations in our Clothing–SMPL registrations are typically minor (e.g., low IR with shallow contacts), such small geometric imperfections, which are common in commercial garment assets and datasets like CLOTH3D \cite{bertiche2020cloth3d}, can still prevent IPC solvers from initializing properly. In contrast, our current simulator is more tolerant and can robustly resolve these cases. Supporting the other solvers would require additional preprocessing to eliminate initial intersections.}

\revised{As future work, several directions could further improve the system. Since the underlying simulator \cite{grigorev2024contourcraft} supports vertex pinning to preserve artist-intended configurations (e.g., keeping a hood up), incorporating optional user-defined pin constraints into our framework can offer additional controllability when desired.} In addition, the current pipeline relies on an intermediate SMPL proxy; a promising direction is to leverage this framework to build large-scale paired data and learn direct mappings from arbitrary garments to arbitrary characters. More broadly, we view our system as a step toward scalable generative apparel digitization, design, and editing.

\section{Conclusion}
In this paper, we introduced a fully automated virtual try-on framework that transfers complex, multilayer garments to a wide range of humanoid models, including realistic humans and stylized characters. By using SMPL as an intermediate proxy, our system generalizes beyond narrowly parameterized settings while avoiding manual intervention.

We decouple clothing-to-body transfer into two alignment problems: clothing-SMPL (partial‐to‐complete alignment) and body-SMPL (large shape/pose variation), and solve each with a correspondence strategy suited to its geometric and appearance variability. Combined with registration and simulation-based fitting, this yields robust draping results across diverse body shapes, garment types, and poses, and generalizes to unseen inputs without retraining.

\begin{acks}
This work was supported by the Metaverse Center Grant from the MBZUAI Research Office.
\end{acks}

\bibliographystyle{ACM-Reference-Format}
\bibliography{sample-base}



\end{document}